\documentclass[10pt,twocolumn,letterpaper]{article}

\usepackage[pagenumbers]{cvpr} % To force page numbers, e.g. for an arXiv version

\usepackage[T1]{fontenc}
\usepackage{cuted} %< for strip environment

\usepackage{currfile} %< for \currfilebase macro

\usepackage{caption} %< for "\captionof" commands (teaser)
\usepackage{amsmath,amssymb,booktabs,xspace,tikz}
\usetikzlibrary{arrows.meta,positioning,fit,calc}
\usepackage[ruled,vlined,linesnumbered]{algorithm2e}
\newcommand{\method}{ConserveEdit\xspace}

\usepackage{amsthm}
\newtheorem{proposition}{Proposition}
\makeatletter
\newcommand{\suppref}[2]{\@ifundefined{r@#1}{#2}{\cref{#1}}}
\makeatother
\theoremstyle{remark}
\usepackage{pifont}
\definecolor{cP}{HTML}{1F5FA8}   % ConserveEdit / contract core
\definecolor{cA}{HTML}{2E7D32}   % accepted / committed
\definecolor{cR}{HTML}{C0392B}   % rejected / violation
\definecolor{cG}{HTML}{6B7B7C}   % neutral
\definecolor{cH}{HTML}{E0A030}   % halo
\definecolor{cPanel}{HTML}{F4F6F9}

\definecolor{cvprblue}{rgb}{0.21,0.49,0.74}
\usepackage[breaklinks,colorlinks,allcolors=cvprblue]{hyperref}

\hypersetup{
	pdftitle={ConserveEdit: Source-Anchored Transactions for Drift-Bounded Iterative Image Editing},
	pdfauthor={Shuyang Jiang, Fucheng Deng, Yuchuan Luo, Zhenyu Wu}
}

\def\confName{CVPR}
\def\confYear{2027}

\title{ConserveEdit: Source-Anchored Transactions for Drift-Bounded\\Iterative Image Editing}

\author{%
	{\normalsize
		\begin{tabular}{@{}l@{\hspace{0.04\textwidth}}l@{}}
			\textbf{Shuyang Jiang}
			&
			\textbf{Fucheng Deng}
			\\
			University of California, Los Angeles
			&
			Aimakj
			\\
			\texttt{shuyangjiang@ucla.edu}
			&
			\texttt{dengfucheng@aimakj.com}
			\\[0.8em]
			\textbf{Yuchuan Luo}
			&
			\textbf{Zhenyu Wu}\textsuperscript{*}
			\\
			College of Computer Science and
			&
			Key Laboratory of Advanced
			\\
			Technology
			&
			Microprocessor Chips and Systems
			\\
			National University of Defense Technology
			&
			College of Computer Science and
			\\
			\texttt{luoyuchuan09@nudt.edu.cn}
			&
			Technology
			\\
			&
			National University of Defense Technology
			\\
			&
			\texttt{wuzhenyu@nudt.edu.cn}
		\end{tabular}%
	}
}

\begin{document}
% Teaser (Fig. 1). Left: code/experiments/make_teaser.py (+ gallery_extract.py) -> fig/teaser_image.pdf and fig/teaser_info.json;
% its right 131 bp (the old aggregate panel) are trimmed here. Right: code/experiments/make_teaser_entities.py ->
% fig/teaser_entities.pdf, read from the 'auto' rows of fig/table_entities.tex (Tab. 3).
% Turns: two of the 19 supplementary-gallery turns on which ConserveEdit commits, chosen as clear examples (stated in the caption).
\twocolumn[{%
\renewcommand\twocolumn[1][]{#1}%
\maketitle
\begin{center}
\captionsetup{type=figure}
\includegraphics[width=0.7354\textwidth,trim=0 0 131bp 0,clip]{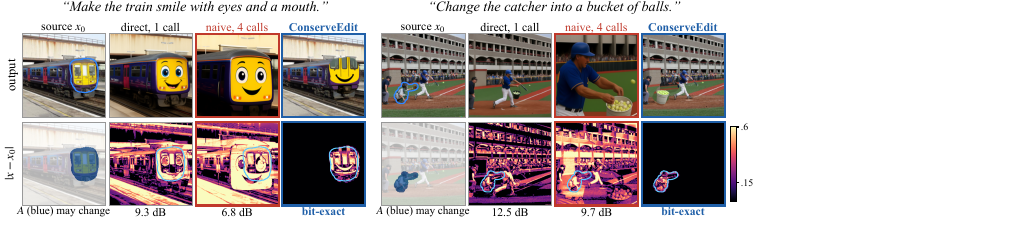}%
\includegraphics[width=0.2646\textwidth]{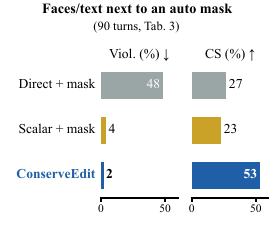}
\captionof{figure}{\textbf{Iterating a strong editor accumulates drift; source-anchored transactions do not.} Left: two MagicBrush test turns with a frozen Qwen-Image-Edit, outputs (top; blue: edit mask) and per-pixel change $|x-x_0|$ with the PSNR outside the dilated mask (bottom; changes below 0.15 from resampling the gallery renders are hidden). One direct call already re-renders the scene, and the naive loop zooms in and repaints it. \method commits only masked candidates that pass separate progress, drift, and quality tests, so everything outside the authorized region $A$ stays bit-identical to $x_0$. Right: next to an automatic mask (\cref{tab:entities}), masking alone changes a protected face or text on 48\% of turns, and masked scalar selection avoids this mostly by abstaining; the separate tests of \method keep violations at 2\% and raise constrained success (CS) to 53\%. Turns: two of the 19 committed turns of the supplementary gallery, chosen as clear examples (their seams are typical, \cref{sec:limits}); \cref{fig:realqual} and the gallery are uncurated.}
\label{fig:teaser}
\end{center}%
}]

\begingroup
\renewcommand{\thefootnote}{\fnsymbol{footnote}}
\footnotetext[1]{Corresponding author.}
\endgroup

\begin{abstract}
Instruction-driven image editors produce convincing local edits in one pass, but inside an iterative refinement loop every accepted output becomes the next input, so unintended changes outside the requested region accumulate.
We present \method, an editor-agnostic controller that treats each refinement step as a transaction against a contract compiled once from the source image.
The contract separates what may change (the authorized region of each target, with a bounded halo for shadows and reflections) from what must survive (protected content with explicit tolerances).
Candidates from a frozen editor are composed through a source-anchored mask and committed only if target progress, drift on protected content, and image quality pass as separate tests; otherwise the step is rolled back and repaired.
For any editor, this keeps every pixel outside the authorized region bit-exact and bounds drift independently of the number of iterations, whereas no fixed scalar trade-off reproduces the separate tests.
On all 1{,}053 MagicBrush test edits with a frozen Qwen-Image-Edit, masking is necessary but not sufficient. A direct call already alters the rest of the image (12.1\,dB PSNR outside the mask); composed through the source-anchored mask, it stays bit-exact there and its DINO similarity to the ground truth rises from 0.67 to 0.85, matching \method, so this gain comes from the mask.
Next to automatic masks, however, masked direct editing changes a protected face or text on 48\% of turns, and masked scalar selection avoids this mostly by abstaining (23\% constrained success). Because its drift test names the violated entity, \method shrinks the support instead (2\% violations, 53\% constrained success); in a simulator with known ground truth it raises constrained success from 55.0\% to 61.8\%.
Automatic grounding remains the main bottleneck.
Code and logs will be released.
\end{abstract}

\section{Introduction}
\label{sec:intro}

% Overview figure: a real MagicBrush test turn (115037_1) with real Qwen-Image-Edit candidates.
% Images and values: code/experiments/fig1_assets.py -> fig/fig1/{*.png,info.json}; ledger from runs/qwen/115037_1/log.json.
\begin{figure*}[t]
\centering
\newcommand{\thumb}[1]{\includegraphics[width=2.05cm]{fig/fig1/#1}}
\resizebox{\textwidth}{!}{%
\begin{tikzpicture}[font=\small, >={Stealth[length=2.2mm]},
  img/.style={inner sep=0pt, draw=black!30, line width=0.4pt},
  plab/.style={font=\small\bfseries, anchor=west, inner sep=1pt, text=black!85},
  capt/.style={font=\footnotesize, align=center, inner sep=1.5pt, anchor=north},
  pnl/.style={draw=black!10, fill=cPanel, rounded corners=4pt},
  mod/.style={draw=cP!70, rounded corners=3pt, fill=white, align=center, font=\footnotesize, inner sep=3pt, line width=0.6pt},
  arr/.style={->, line width=0.8pt, black!55}]
% panels
\fill[pnl] (-0.2,-2.5) rectangle (5.0,1.9);
\fill[pnl] (5.25,-2.5) rectangle (14.55,1.9);
\fill[pnl] (14.8,-2.5) rectangle (21.55,1.9);
\node[plab] at (-0.1,1.62) {\textcolor{cP}{\large\textbf{1}}\; Compile contract once from $x_0$};
\node[plab] at (5.35,1.62) {\textcolor{cP}{\large\textbf{2}}\; Frozen editor proposes $B$ candidates; audit vs.\ $x_0$};
\node[anchor=west, font=\footnotesize, text=black!50] at (5.38,1.3) {raw outputs re-render the whole image (PSNR$_{out}$ 8.6--9.5\,dB; 3 of 4 shown)};
\node[plab] at (14.9,1.62) {\textcolor{cP}{\large\textbf{3}}\; Masked commit or roll back};
% panel 1
\node[img] (src) at (1.15,0.05) {\thumb{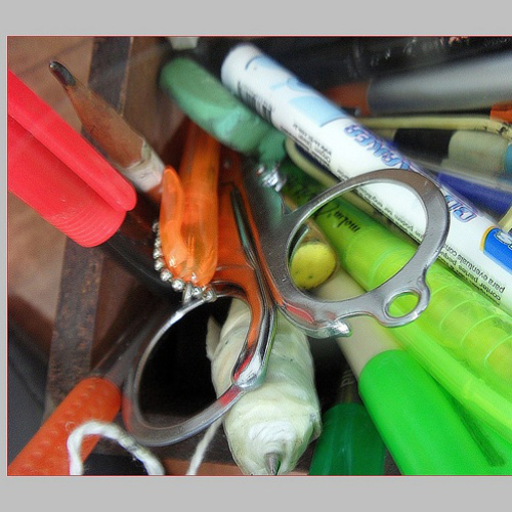}};
\node[capt] at (src.south) {source $x_0$\\ ``Let the scissors be\\ made of plastic.''};
\node[img] (con) at (3.75,0.05) {\thumb{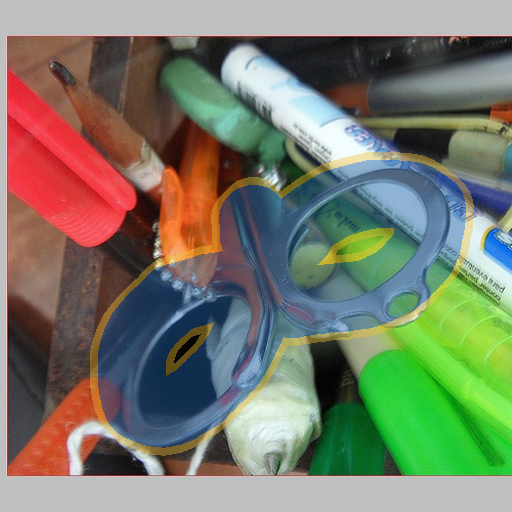}};
\node[capt] at (con.south) {\textcolor{cP}{$\blacksquare$} core $M$ (human)\\ \textcolor{cH}{$\blacksquare$} halo $H$ (8\,px)\\ rest: protected};
\draw[arr] (src) -- (con);
% panel 2
\node[mod] (ex) at (5.95,0.05) {Qwen-\\Image-\\Edit\\ (frozen)};
\draw[arr] (con.east) -- (ex.west);
\node[img] (c0) at (7.95,0.05) {\thumb{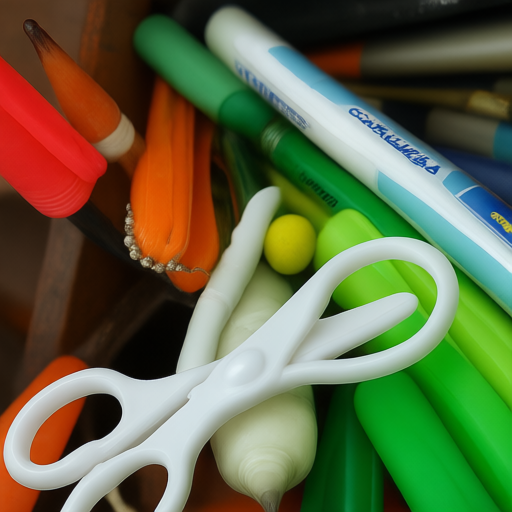}};
\node[img] (c3) at (10.45,0.05) {\thumb{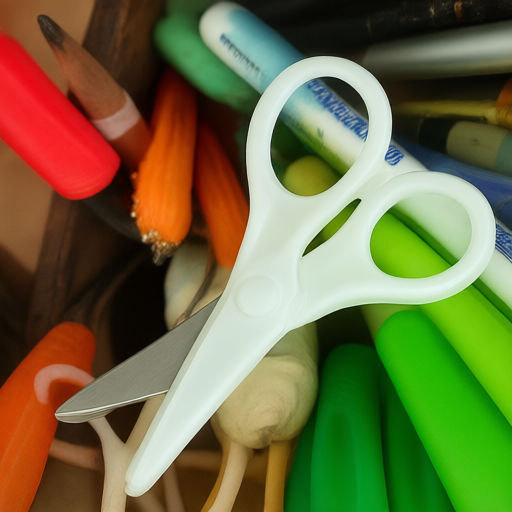}};
\node[img, draw=cA, line width=1.2pt] (c2) at (12.95,0.05) {\thumb{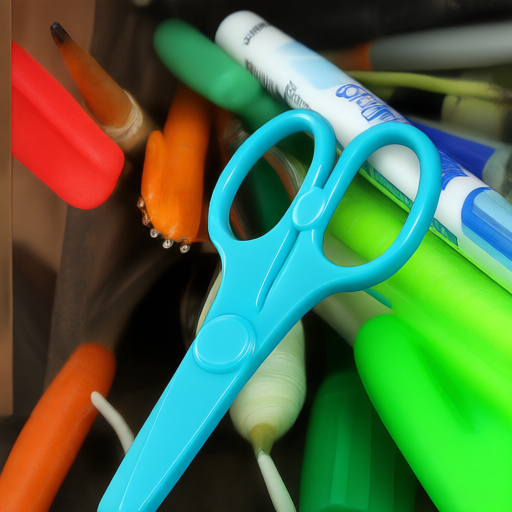}};
\draw[arr] (ex.east) -- (c0.west);
\node[capt] (k0) at (c0.south) {$\Delta f{=}0.00{<}\delta$\\ \textcolor{cR}{\ding{55}\,\textbf{no progress}}};
\node[capt] at (c3.south) {$\Delta f{=}0.20$\\ \textcolor{cG}{\ding{51}\,feasible}\\ \textcolor{black!50}{not selected}};
\node[capt] at (c2.south) {$\Delta f{=}0.35$ (largest)\\ \textcolor{cA}{\ding{51}\,\textbf{feasible}}\\ \textcolor{cA}{\textbf{selected}}};
% panel 3
\node[img, draw=cA, line width=1.2pt] (out) at (16.1,0.05) {\thumb{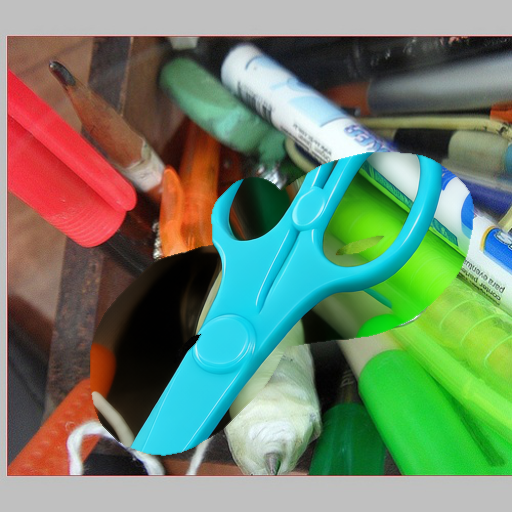}};
\node[capt] at (out.south) {committed $x_1$\\ outside $A$: \textbf{bit-exact}};
\draw[arr, cA, line width=1.1pt] (c2.east) -- (out.west);
\node[mod, anchor=west, align=left] (led) at (17.4,0.05) {%
\textbf{\textcolor{cP}{Decision log}}\\[2pt]
step 1: 2 rejected (progress)\\
\quad 2 feasible $\rightarrow$ commit\\
editor calls: 4};
\draw[arr] (out.east) -- (led.west);
\draw[arr, cR, dashed] (k0.south) -- (k0.south |- 0,-2.02) -| (led.south);
\node[font=\footnotesize, text=cR, anchor=north] at (10.45,-2.04) {rejected candidates are rolled back and logged};
\end{tikzpicture}}
\caption{\textbf{\method on a real MagicBrush test turn with a frozen Qwen-Image-Edit.} \textbf{1}~The contract is compiled once from $x_0$; the region outside the authorized support $A{=}M\cup H$ is protected. \textbf{2}~Every raw candidate changes the whole image, so a scalar loop would have to trade that drift against progress. Each candidate is instead composed through the commit mask and audited for progress and quality as separate conditions. \textbf{3}~The feasible candidate with the largest progress is committed and every decision is logged. Shown is the first step of the run; all images and numbers come from the released code (\cref{sec:real}).}
\label{fig:overview}
\end{figure*}

Latent diffusion and flow-matching generators~\cite{rombach2022ldm,lipman2023flow,esser2024scaling} have turned instruction-based image editing into a general interface: Step1X-Edit~\cite{liu2025step1x}, In-Context Edit~\cite{zhang2025icedit}, FLUX.1 Kontext~\cite{labs2025kontext}, and Qwen-Image-Edit~\cite{wu2025qwenimage} accept a source image and free-form text and return a realistic edit in one pass.
A strong one-step editor does not, however, define a reliable \emph{multi-step} process: each new generation may alter the requested target and, at the same time, content the user never asked to touch.

Agentic editors wrap the editor in a perceive--reason--act loop~\cite{wang2024genartist,li2025editthinker,zeng2026mira,yin2025reasonedit,ji2025planning}, and their acceptance step usually maximizes one aggregate quality signal.
An aggregate permits \emph{compensation}: a large gain on the requested object can hide a small identity shift, corrupted text, or a displaced background.
Because the next iteration consumes the accepted image, collateral changes become persistent state and accumulate.

We adopt a stricter premise: \emph{content outside the authorized causal support of an edit is an invariant, not a soft preference.}
Evaluation studies support judging instruction following, quality, and preservation separately~\cite{xu2025lmm4edit,gao2026tiedit,liu2026judges,long2026spatialreward}. Unlike consistency control inside one generator~\cite{yin2025consistedit,cui2026samflow,kang2026inline}, we seek a controller outside the editor that makes an auditable decision at every step.

\method realizes this premise as a transactional loop around a frozen executor (\cref{fig:overview}).
A request is compiled once into a persistent \emph{edit contract}. Each iteration proposes atomic candidates, composes them through a source-anchored mask, audits progress and drift as separate conditions, and commits only a feasible candidate; otherwise it rolls back and logs a repair.

\noindent Our contributions are:
\begin{itemize}
\item a persistent edit contract, compiled once from the source, that separates what may change from what must survive (\cref{sec:editcontract});
\item a transactional refinement loop that commits a masked candidate only if progress, drift, and quality pass as separate tests, with source invariance and an iteration-independent drift bound for any executor, and a proof that no scalar weight reproduces these tests (\cref{sec:transactionalrefinement,sec:theory});
\item a real-editor study with a frozen Qwen-Image-Edit on all 1{,}053 MagicBrush test turns (\cref{sec:real}) showing that masking is necessary but not sufficient: one masked call already keeps the unedited region bit-exact and raises DINO similarity to the ground truth from 0.67 to 0.85, but next to automatic masks only the separate tests protect faces and text (violations 2\% vs.\ 48\%) without abstaining on most turns (constrained success 53\% vs.\ 23\% for masked scalar selection). With only the background protected, the gate matches masked scalar selection;
\item a simulation with known ground truth that isolates the acceptance logic and identifies grounding as its bottleneck (\cref{sec:sim}).
\end{itemize}

\section{Related Work}
\label{sec:related}

\subsection{Diffusion-based image editing}
Diffusion and flow generators~\cite{ho2020ddpm,song2021ddim,ho2022cfg,peebles2023dit,liu2023rectified} enabled zero-shot editing by noising and denoising~\cite{meng2022sdedit}, attention control~\cite{hertz2023p2p,tumanyan2023pnp,cao2023masactrl}, inversion~\cite{mokady2023nti,ju2024pnpinversion}, and optimization of text embeddings~\cite{kawar2023imagic}.
Mask-based methods confine changes spatially, by blending~\cite{avrahami2022blended,avrahami2023bld}, automatic mask inference~\cite{couairon2023diffedit}, or dedicated inpainting~\cite{lugmayr2022repaint,xie2023smartbrush,wang2023imageneditor,ju2024brushnet,zhuang2024powerpaint,yang2023pbe}.
Our commit mask is closest in spirit to blending, but it acts on \emph{completed} candidates across a trajectory rather than inside one denoising run; we compare against mask-conditioned inpainting in \cref{sec:real}.

\subsection{Instruction-based editing}
InstructPix2Pix learned language-driven editing from synthetic triplets~\cite{brooks2023instructpix2pix}, and MagicBrush contributed human-annotated multi-turn data~\cite{zhang2023magicbrush}.
Later work scaled data and tasks~\cite{zhao2024ultraedit,hui2024hqedit,yu2025anyedit,ye2025imgedit,geng2023instructdiffusion,sheynin2024emuedit} and added MLLM guidance~\cite{fu2024mgie,huang2024smartedit}.
Recent unified systems scale the generator and its conditioning~\cite{liu2025step1x,zhang2025icedit,labs2025kontext,wu2025qwenimage,xiao2025omnigen}.
We use such models as frozen executors or direct baselines.

\subsection{Spatial and consistency control}
ControlNet and IP-Adapter add structural or image-prompt conditions~\cite{zhang2023controlnet,ye2023ipadapter}; ConsistEdit manipulates MM-DiT attention~\cite{yin2025consistedit}; SAM-Flow anchors a masked flow trajectory to the source~\cite{cui2026samflow}; Inline Critic steers hidden states with a critic token~\cite{kang2026inline}.
\method alters neither attention, latents, nor weights. It governs which completed candidate may become persistent state, which makes it complementary to these controls.

\subsection{Reasoning and agentic editing}
LLM agents interleave reasoning, tool use, and self-feedback~\cite{yao2023react,madaan2023selfrefine,shinn2023reflexion}; Visual ChatGPT and GenArtist bring this to visual tools~\cite{wu2023visualchatgpt,wang2024genartist}.
ReasonEdit adds reflection~\cite{yin2025reasonedit}; RePlan grounds instructions in regions~\cite{qu2026replan}; EditThinker critiques outputs and rewrites prompts~\cite{li2025editthinker}; MIRA predicts atomic actions~\cite{zeng2026mira}; PhotoAgent plans aesthetic edits by tree search~\cite{yao2026photoagent}; planning-based editors decompose requests~\cite{ji2025planning}.
Closest to our goal, Agent Banana observes that non-edited regions drift toward the generator's prior under repeated re-generation and confines each edit to a cropped image layer~\cite{ye2026agentbanana}. Like our commit mask, this bounds where pixels may change. \method adds what spatial confinement alone does not provide: a contract anchored to the source, acceptance tests that cannot trade drift for progress, and a logged decision for every candidate, with guarantees that hold for any executor.
EditRefiner iterates perception, reasoning, action, and evaluation but optimizes post-edit quality rather than source invariants~\cite{xu2026editrefiner}.
We compare against EditThinker, RePlan, and EditRefiner with their released weights (\cref{sec:real}); MIRA and Agent Banana have no released inference code.
ImageEdit-R1 and SpatialReward improve policies or rewards by RL~\cite{zhao2026imageeditr1,long2026spatialreward}, yet a better scalar evaluator does not by itself prevent compensation.
\method makes these dimensions non-substitutable at acceptance and separates generation from commitment.

\subsection{Editing evaluation}
CLIP-based scores~\cite{radford2021clip,hessel2021clipscore,gal2022nada} and perceptual or self-supervised features~\cite{zhang2018lpips,oquab2024dinov2} are standard; VIEScore and LMM4Edit use MLLM evaluators~\cite{ku2024viescore,xu2025lmm4edit}, whose reliability as judges is itself debated~\cite{zheng2023judge,liu2026judges}. TIEdit shows that common metrics correlate imperfectly with expert preservation scores~\cite{gao2026tiedit}.
We report each dimension separately with an evaluator disjoint from the controller.

\section{Method}
\label{sec:method}

\subsection{Problem formulation and overview}
Let $x_0\in[0,1]^{H\times W\times 3}$ be a source image and $u$ a free-form instruction.
A conventional iterative editor applies $x_{t+1}=E_\phi(x_t,p_t,z_t)$ with a refined prompt $p_t$ and seed $z_t$.
Because every accepted output is the next input, preservation error can accumulate even when each step looks locally beneficial.
We seek to maximize satisfaction of $u$ subject to explicit bounds on changes to content that $u$ does not authorize.

We distinguish a semantic target from its \emph{causal support}.
Replacing a lamp authorizes the lamp pixels and bounded changes to its shadow or reflection; it does not authorize changes to a face, a caption, or a distant object.
A target mask is thus not an exact isolation boundary.
\method represents each operation by a core support plus an uncertainty-aware halo and audits all remaining content against the immutable source.

\method has two components: a persistent contract compiled once from $(x_0,u)$ (\cref{sec:editcontract}), and a transactional loop that commits masked candidates only when they pass separate feasibility tests (\cref{sec:transactionalrefinement}). \Cref{sec:theory} states what this guarantees; \Cref{fig:overview} and \cref{alg:loop} summarize the loop.

\subsection{Persistent edit contract}
\label{sec:editcontract}
A contract compiler runs once on $(x_0,u)$ and outputs
\begin{equation}
\mathcal{C}=\big(G,\;A,\;P,\;\epsilon,\;\tau\big),
\label{eq:contract}
\end{equation}
where $G=\{g_j\}_{j=1}^{J}$ are target clauses, $A=\{A_j\}_{j=1}^{J}$ their authorized supports, $P=\{P_k\}_{k=1}^{K}$ protected entities, and $\epsilon,\tau$ the preservation tolerances and clause thresholds.
The contract stays anchored to $x_0$: later observations may refine uncertain geometry but may not delete a protected entity or lower a threshold without an explicit, logged revision.

\paragraph{Target clauses.}
Each clause $g_j=(o_j,v_j,r_j,\pi_j)$ specifies an operation, a visual subject, a relational context, and an executable predicate $\pi_j$ that maps a source--candidate pair to a score $f_j(x_0,x)\in[0,1]$.
Predicates can be specialized per operation (\eg, OCR exact match for text edits), which reduces reliance on any single MLLM judgment; our experiments use one local CLIP directional score (\cref{sec:real}).

\paragraph{Authorized support.}
$A_j=M_j\cup H_j$ joins a core mask $M_j$ with a bounded causal halo $H_j$ for shadows, reflections, or local illumination.
A user mask, when given, is $M_j$; otherwise Grounding DINO~\cite{liu2024groundingdino} detects a phrase taken from the instruction by fixed rules, and SAM~\cite{kirillov2023sam} segments it.
If the masks of four boxes jittered by $\pm5\%$ disagree (IoU${<}\tau_{env}$, calibrated on the dev split), the controller abstains instead of issuing a preservation claim, treating grounding as a measured uncertainty rather than an oracle.

\paragraph{Protected entities.}
For every $P_k$ the compiler stores source descriptors suited to its content, such as pixel and perceptual features~\cite{zhang2018lpips}, semantic embeddings~\cite{radford2021clip,oquab2024dinov2}, face identity~\cite{deng2019arcface}, or OCR tokens, so that a person and a printed sign receive different tests.
Tolerances $\epsilon_k$ are calibrated on source-preserving transformations and frozen before test evaluation. Our experiments use pixel descriptors for the background (plus edges in simulation) and, on real images, ArcFace identity for faces and OCR strings plus pixel MAE for text (\cref{sec:real}).

The persistent contract prevents \emph{goalpost drift}: a controller that re-describes the current image at every turn can gradually accept an altered identity because the altered state becomes its new reference.
In \method the current image supplies evidence about progress, but protected descriptors and tolerances always originate from $x_0$.

\subsection{Transactional refinement, complement audit and feasible acceptance}
\label{sec:transactionalrefinement}
\paragraph{Masked transactions.}
At iteration $t$ the state is
\begin{equation}
s_t=(x_0,\,x_t,\,u,\,\mathcal{C},\,U_t,\,\mathcal{L}_t),
\end{equation}
with cumulative authorized support $U_t$ and decision log $\mathcal{L}_t$.
The controller selects an unsatisfied clause and proposes $B$ actions $a_{t,b}=(j,p_{t,b},W_{t,b},z_{t,b})$: an atomic instruction, a feathered commit mask $W_{t,b}\in[0,1]^{H\times W}$ contained in $A_j$, and a seed.
The frozen executor generates
\begin{equation}
\tilde{x}_{t,b}=E_\phi(x_t,p_{t,b},z_{t,b}),
\label{eq:gen}
\end{equation}
and the proposed transaction image is the masked composition
\begin{equation}
x^{+}_{t,b}=W_{t,b}\odot\tilde{x}_{t,b}+(1-W_{t,b})\odot x_t.
\label{eq:compose}
\end{equation}
Because every committed update is masked, pixels outside $U_t=\bigcup_{\ell\le t}A_{j_\ell}$ equal $x_0$ \emph{by construction}.
This operational guarantee covers only the represented support; it does not imply semantic correctness when a mask omits a true causal dependency, which is why halo expansion is a first-class, logged repair action.

\paragraph{Complement audit.}
For protected entity $P_k$ with valid descriptor set $V_k$, descriptor extractors $F_m$, and distances $d_m$, the drift is
\begin{equation}
D_k(x_0,x)=\sum_{m\in V_k} w_{k,m}\, d_m\!\big(F_m(P_k\odot x_0),\,F_m(P_k\odot x)\big).
\label{eq:drift}
\end{equation}
A dense change map additionally flags differences that escape $U_t$, and a perceptual quality score $q(x)$ is computed independently so that compositing cannot buy preservation with visible seams.

\paragraph{Feasible acceptance and repair.}
A candidate is \emph{feasible} iff
\begin{align}
D_k(x_0,x^{+}_{t,b}) &\le \epsilon_k \quad \forall k, \label{eq:c1}\\
f_j(x_0,x^{+}_{t,b})-f_j(x_0,x_t) &\ge \delta_j, \label{eq:c2}\\
q(x^{+}_{t,b}) &\ge q_{\min}. \label{eq:c3}
\end{align}
\Cref{eq:c1} makes preservation a hard gate, \cref{eq:c2} rejects harmless but ineffective edits, and \cref{eq:c3} rejects visibly degraded transactions.
Among feasible candidates, \method selects lexicographically by minimum normalized clause gain, then quality, then cost; a weighted scalar only breaks remaining ties.

When no candidate is feasible the controller rolls back and applies a contract-preserving repair: shrink the core mask, expand only the halo, or resample the seed (suppl.\ \suppref{sup:impl}{Sec.~A} gives the order).
It cannot keep the failed image as hidden state.
Each rejection logs the violated clause, measured drift, and repair, enabling analysis of false rejection, false acceptance, and rollback frequency; repeated action signatures that already failed under an unchanged contract are disallowed to prevent cycling.

\begin{algorithm}[t]
\caption{Transactional refinement in \method}
\label{alg:loop}
\small
\KwIn{source $x_0$, instruction $u$, editor $E_\phi$, budget $T$, width $B$}
Compile $\mathcal{C}$ from $(x_0,u)$; $x\leftarrow x_0$; $\mathcal{L}\leftarrow\emptyset$\;
\For{$t=0,\dots,T-1$}{
  choose unsatisfied clause $j$; propose $\{a_{t,b}\}_{b=1}^{B}$\;
  \ForEach{$a_{t,b}$}{
    generate $\tilde{x}_{t,b}$ (\cref{eq:gen}); compose $x^{+}_{t,b}$ (\cref{eq:compose})\;
    audit progress, drift, escaped change, quality\;
  }
  \eIf{some candidate satisfies \cref{eq:c1,eq:c2,eq:c3}}{
    commit lexicographic best; record the decision in $\mathcal{L}$\;
  }{
    roll back; log violations; apply repair\;
  }
  \lIf{all clauses pass under two audits \textbf{or} no repair remains}{\textbf{break}}
}
\Return{last committed image and $\mathcal{L}$}
\end{algorithm}

\paragraph{Stopping and logging.}
The loop stops successfully when every clause exceeds $\tau_j$, every entity stays below $\epsilon_k$, and the decision is reproduced by a second audit with perturbed evaluator order.
It stops conservatively when the budget is exhausted, no admissible repair remains, or grounding uncertainty invalidates the contract.
The returned log lists clause scores, drift values, authorized regions, editor calls, rejected transactions, and the stop reason.
These make the drift bound auditable; human evaluation is still needed to test whether the contract captured user intent.

\subsection{Properties}
\label{sec:theory}
Let $\mathrm{supp}(W_t)\subseteq A_{j_t}$ for every committed step. The following properties hold for any, even adversarial, executor $E_\phi$; proofs are in suppl.\ \suppref{sup:theory}{Sec.~D}.

\begin{proposition}[Source invariance and iteration-independent drift]\label{prop:inv}
For every $t$ and every pixel $p\notin U_t$, $x_t(p)=x_0(p)$ exactly. Moreover, every committed state satisfies $D_k(x_0,x_t)\le\epsilon_k$ for all $k$, so drift is bounded independently of $T$ and $B$.
\end{proposition}

\begin{proposition}[No scalar weight recovers the feasible set]\label{prop:scalar}
Let $J=f-\lambda D$ with a fixed $\lambda>0$, and commit $x^+$ iff $J(x^+)>J(x_t)$. If $D(x_0,x_t)=0$ and $0<\delta<1-f(x_t)$, no $\lambda$ accepts exactly the candidates satisfying \cref{eq:c1,eq:c2}: for $\lambda<(1-f(x_t))/\epsilon$ some infeasible candidate is accepted, and for $\lambda\ge\delta/\epsilon$ some feasible candidate is rejected.
\end{proposition}

If the audit instead compared each candidate with the current image $x_t$, drift could grow linearly in $T$ (suppl.\ \suppref{prop:goal}{Proposition~S1}). Both properties concern the represented contract: if grounding misses part of the target, invariance still holds but the edit may be incomplete, which \cref{sec:sim} measures.

\section{Experiments}
\label{sec:exp}

We first evaluate \method with a frozen real editor on MagicBrush (\cref{sec:real}) and then analyze its acceptance logic in a simulator with known ground truth (\cref{sec:sim}).

\subsection{Real-Editor Study: Qwen-Image-Edit on MagicBrush}
\label{sec:real}

\paragraph{Setup.} \emph{Data:} all 1{,}053 turns (535 sessions) of the MagicBrush test split~\cite{zhang2023magicbrush}; each has a source image, an instruction, a human-drawn edit mask, and a ground-truth (GT) target, all resized to $512^2$.
\emph{Executor:} the official Qwen-Image-Edit checkpoint~\cite{wu2025qwenimage} (bf16, 20 steps, true CFG 4.0), identical for every method; calls with identical (input, prompt, seed) are generated once and shared across controllers.
\emph{Contract:} the human mask is the core support, with an 8\,px halo, and the rest of the image is a protected background entity (pixel MAE, $\epsilon{=}0.02$). The target predicate is a local CLIP ViT-B/32 directional score in the mask's bounding box ($\tau{=}0.5$, $\delta{=}0.02$); the budget is $T{=}2$, $B{=}4$.
\emph{Baselines:} direct editing, the naive loop, best-of-$B$ scalar selection, and the unmasked gate let the editor change the whole image. Direct\,+\,mask composes the single direct output through the commit mask of \method (one call, no test); FLUX.1 Fill [dev]~\cite{bfl2024fill} inpaints the same support $A$ and is composed through the same $W$; Scalar\,+\,mask selects among the same masked candidates by $f-\lambda D$.
\emph{Evaluator:} disjoint from the controller, it reports L1, CLIP-I (ViT-L/14), and DINOv2 similarity to the GT, the CLIP ViT-L/14 directional score CLIP-D between image change and instruction, and PSNR outside the dilated mask. Tests are paired sign-flip permutations, Holm-corrected over the four GT and CLIP metrics where stated.

\begin{table*}[t]
\centering
\caption{\textbf{Real editor: Qwen-Image-Edit on all 1{,}053 MagicBrush test turns.} Abstain: output equals $x_0$. $^\dagger$Never commits, so its GT similarities (in parentheses) are those of the unedited source. $^\ddagger$Receives the support $A$ as its inpainting mask and the instruction as its prompt. ``exact'': outside-mask pixels are bit-identical to $x_0$. Calls: editor calls. Published agentic editors: suppl.\ \suppref{tab:s_agents}{Tab.~S2}; detector predicates: suppl.\ \suppref{tab:s_pred}{Tab.~S6}.}
\label{tab:real}
\small
\setlength{\tabcolsep}{8pt}
% Generated by code/experiments/make_table_real.py from results_v2/qwen_full/summary.json + results_v2/fill_human/summary.json (n=1053 MagicBrush test turns).
% The agentic rows moved to fig/suppl/table_agents.tex and the detector-predicate row to Tab. S6 (fig/suppl/table_pred.tex).
\begin{tabular}{@{}lccccccc@{}}
\toprule
Controller & CLIP-D$\uparrow$ & Abstain & L1$_{gt}\!\downarrow$ & CLIP-I$_{gt}\!\uparrow$ & DINO$_{gt}\!\uparrow$ & PSNR$_{out}\!\uparrow$ & Calls \\
\midrule
\multicolumn{8}{@{}l}{\emph{Editor changes the whole image}} \\
Direct (1 call) & 0.058 & 0\% & 0.196 & 0.820 & 0.666 & 12.1 & 1.0 \\
Naive loop & 0.058 & 0\% & 0.262 & 0.734 & 0.485 & 9.9 & 3.4 \\
Best-of-$B$ scalar & 0.057 & 27\% & 0.149 & 0.849 & 0.719 & 25.2 & 6.9 \\
Gate, no mask$^\dagger$ & 0.000 & 100\% & (0.036) & (0.930) & (0.868) & exact & 8.0 \\
\midrule
\multicolumn{8}{@{}l}{\emph{Committed through the human mask}} \\
Direct\,+\,mask (1 call) & 0.069 & 0\% & 0.047 & 0.896 & 0.853 & exact & 1.0 \\
FLUX.1 Fill, same mask$^\ddagger$ & 0.064 & 0\% & \textbf{0.037} & \textbf{0.930} & \textbf{0.870} & exact & 1.0 \\
Scalar\,+\,mask & \textbf{0.083} & 12\% & 0.047 & 0.894 & 0.849 & exact & 6.8 \\
\method & 0.081 & 13\% & 0.047 & 0.896 & 0.850 & exact & 6.9 \\
\bottomrule
\end{tabular}

\end{table*}

\begin{figure*}[t]
\centering
\includegraphics[width=\textwidth]{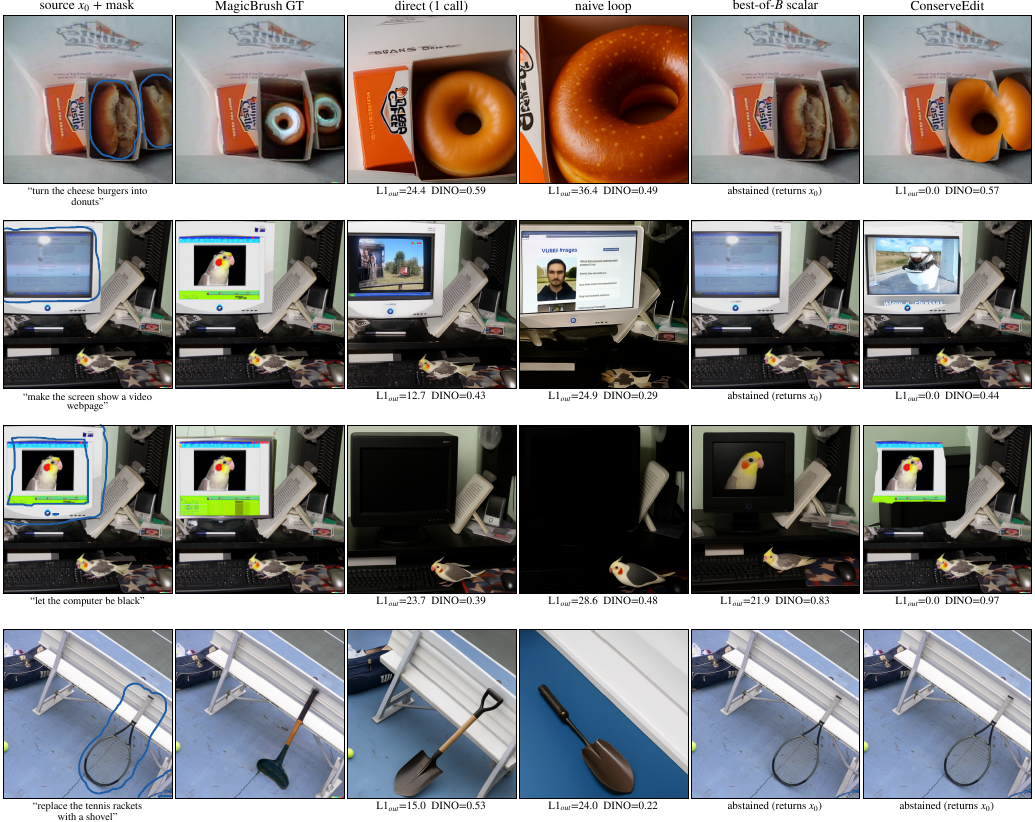}
\caption{\textbf{Real Qwen-Image-Edit outputs on MagicBrush} (first four turns in id order with mask area in $[5\%,40\%]$). Direct editing and the naive loop re-render the whole scene (L1$_{out}$ in $10^{-2}$), and best-of-$B$ scalar selection still drifts or abstains. \method leaves everything outside the mask bit-exact. In row 4 the local CLIP predicate rejects all candidates and \method abstains; rows 1 and 2 show seams at the mask boundary (\cref{sec:limits}).}
\label{fig:realqual}
\end{figure*}

\paragraph{Drift of the real editor.} One Qwen-Image-Edit call already changes the region outside the mask (PSNR$_{out}$ 12.1\,dB; \cref{fig:realqual}). Feeding outputs back (naive loop) compounds this: PSNR$_{out}$ falls to 9.9\,dB and DINO similarity to the GT drops from 0.666 to 0.485. At the editor's preferred $1024^2$ resolution, direct editing drifts less (22.3 vs.\ 12.2\,dB on a 160-turn subset); the comparisons below hold there too, except that \method's CLIP-D advantage over direct editing is no longer significant ($p{=}0.43$; suppl.\ \suppref{tab:s_res}{Tab.~S4}).

\paragraph{Effect of masked commitment.} \method improves on direct editing in every GT metric (L1$_{gt}$ 0.047 vs.\ 0.196, DINO$_{gt}$ 0.850 [0.84,\,0.86] vs.\ 0.666 [0.65,\,0.68]) and in CLIP-D (0.081 vs.\ 0.058), all $p{<}10^{-3}$. Direct\,+\,mask, the same direct output composed through the same mask, matches this GT similarity (DINO$_{gt}$ 0.853, $p{=}0.21$; L1$_{gt}$ $p{=}0.75$): the preservation gain comes from masked commitment, not from iteration. The extra calls raise only CLIP-D (0.081 vs.\ 0.069, $p_{\mathrm{Holm}}{<}10^{-3}$), and masked scalar selection gains as much (0.083), so this gain comes from choosing among candidates. FLUX.1 Fill reaches the highest GT similarity (DINO$_{gt}$ 0.870) but the weakest masked edit (CLIP-D 0.064); since the unedited source also scores 0.868, GT similarity on these turns rewards conservative fills as much as correct edits.

\paragraph{Gate versus masked scalar selection.} With a background-only contract, \method and Scalar\,+\,mask produce identical outputs on 92\% of turns; the remaining differences are small and go both ways (CLIP-D 0.081 vs.\ 0.083, $p{=}0.004$; L1$_{gt}$ 0.0471 vs.\ 0.0474, $p{<}10^{-3}$; DINO$_{gt}$ $p{=}0.16$). Without a mask the gate never commits, because all 8{,}424 unmasked candidates exceed the background tolerance, whereas unmasked scalar selection trades drift for progress (PSNR$_{out}$ 25.2\,dB), as \cref{prop:scalar} predicts. \method abstains on 12.9\% of turns (Scalar\,+\,mask: 11.8\%), mostly because the local CLIP predicate misses valid edits (3{,}793 progress versus 141 seam rejections); dev-calibrated detector predicates trade these abstentions for others (\cref{sec:limits}).

\paragraph{Published agentic editors.} On a 160-turn subset with the same editor and seeds and at most eight editor calls (suppl.\ \suppref{sup:agents}{Sec.~A}), EditThinker~\cite{li2025editthinker}, EditRefiner~\cite{xu2026editrefiner}, and RePlan~\cite{qu2026replan} do not preserve the source by themselves (DINO$_{gt}$ 0.708, 0.697, and 0.837 vs.\ 0.867 for \method, $p_{\mathrm{Holm}}{\le}0.009$). Composed through the same $W$, all three match \method's GT similarity ($p_{\mathrm{Holm}}{\ge}0.07$), and EditThinker and EditRefiner match Direct\,+\,mask in CLIP-D (0.069 and 0.070 vs.\ 0.070): their MLLM feedback adds no measurable gain over one masked call, and preservation again comes from the mask.

% Table 2. Body generated by code/experiments/make_table_auto.py from results_v2/qwen_auto_full/analysis_exp2.json + results_v2/fill/analysis_fill.json (n=1053);
% auto rows: runs_v2/qwen/auto_full; same GPU type, shared step-0 call cache. Human-mask rows are in Tab. 1.
\begin{table}[t]
\centering
\caption{\textbf{Automatic masks} (Grounding DINO\,+\,SAM with envelope abstention, $\tau_{env}{=}0.8$; all 1{,}053 turns). Abstain includes grounding abstentions (33\%). L1$_{out}$ ($10^{-2}$): change outside the dilated human mask. Mean IoU with the human mask: 0.31. Human-mask rows: \cref{tab:real}.}
\label{tab:auto}
\small
\setlength{\tabcolsep}{3pt}
\begin{tabular}{@{}lcccc@{}}
\toprule
Controller & Abstain & CLIP-D$\uparrow$ & DINO$_{gt}\!\uparrow$ & L1$_{out}\!\downarrow$ \\
\midrule
Direct\,+\,mask & 33\% & 0.037 & 0.816 & 2.80 \\
FLUX.1 Fill & 33\% & 0.025 & 0.769 & 3.19 \\
Scalar\,+\,mask & 40\% & 0.046 & 0.811 & 2.77 \\
\method & 41\% & 0.046 & 0.812 & 2.74 \\
\bottomrule
\end{tabular}
\end{table}

\paragraph{Automatic masks.} \Cref{tab:auto} replaces the human mask with the grounding stack of \cref{sec:editcontract} ($\tau_{env}$ calibrated on dev; suppl.\ \suppref{sup:ground}{Sec.~A}). Grounding abstains on 343 of 1{,}053 turns, and the automatic masks overlap the human ones poorly (mean IoU 0.31). Outputs stay bit-exact outside their own support, but abstention rises from 13\% to 41\% and DINO$_{gt}$ falls from 0.850 to 0.812 ($p_{\mathrm{Holm}}{<}10^{-3}$), still well above direct editing (0.666). The loss is confined to poor grounding: at IoU${\ge}0.6$ (139 turns), automatic and human masks are equivalent (DINO$_{gt}$ 0.809 vs.\ 0.807, $p{=}0.70$). \method again matches Direct\,+\,mask in GT similarity ($p{=}0.16$) and exceeds it in CLIP-D ($p_{\mathrm{Holm}}{<}10^{-3}$), and it now beats FLUX.1 Fill on both counts (CLIP-D 0.046 vs.\ 0.025, DINO$_{gt}$ 0.812 vs.\ 0.769, $p_{\mathrm{Holm}}{<}10^{-3}$).

% Body generated by code/experiments/entity_eval.py (results_v2/entities/analysis_entities.json).
\begin{table}[t]
\centering
\caption{\textbf{Protected faces and text} (Qwen-Image-Edit; the 90 MagicBrush test turns with a face or text instance outside the human mask but within 32\,px of the human or automatic support; 93 faces, 28 text instances). $^\ast$Contract with face (ArcFace) and text (OCR\,+\,MAE) descriptors; Scalar\,+\,mask adds them to $D$ ($\lambda{=}1$). Viol.: a separate face recognizer or OCR finds a protected instance changed; CS: committed, CLIP-D${>}0$, and no violation.}
\label{tab:entities}
\small
\setlength{\tabcolsep}{1.7pt}
\begin{tabular}{@{}llccccc@{}}
\toprule
Mask & Controller & Viol.$\downarrow$ & CS$\uparrow$ & Abstain & CLIP-D$\uparrow$ & DINO$_{gt}\!\uparrow$ \\
\midrule
human & Direct\,+\,mask & 1\% & 80\% & 0\% & 0.059 & 0.850 \\
human & Scalar\,+\,mask$^\ast$ & 1\% & 82\% & 11\% & 0.079 & 0.847 \\
human & {\method}$^\ast$ & 0\% & 83\% & 10\% & 0.078 & 0.847 \\
\midrule
auto & Direct\,+\,mask & 48\% & 27\% & 18\% & 0.037 & 0.808 \\
auto & Scalar\,+\,mask$^\ast$ & 4\% & 23\% & 68\% & 0.028 & 0.861 \\
auto & {\method}$^\ast$ & 2\% & 53\% & 32\% & 0.038 & 0.830 \\
\bottomrule
\end{tabular}
\end{table}

\paragraph{Protected faces and text.} The separate tests should matter when protected content lies near or inside the support. We select, without looking at outputs, the 90 test turns with a face (insightface) or text (EasyOCR) instance outside the human mask but within 32\,px of the human or automatic support (93 faces, 28 text instances), add face-identity (ArcFace) and OCR descriptors with tolerances calibrated on dev, and evaluate with a different face recognizer and OCR (suppl.\ \suppref{sup:entities}{Sec.~A}). With human masks the support rarely reaches these entities, and all masked controllers reach 80--83\% constrained success (CS; \cref{tab:entities}). With automatic masks, Direct\,+\,mask changes a protected entity on 48\% of turns. Scalar\,+\,mask, with the same descriptors in its objective, avoids this only by abstaining on 68\% of turns (CS 23\%): its scalar score does not say which entity was violated. \method rejects the violating candidates, and because its drift test names the violated entity, the shrink repair removes that entity from the support (44 of 90 turns). It keeps violations at 2\% while committing on 68\% of turns, and its CS of 53\% is 30 points above Scalar\,+\,mask and 29 points above \method with background-only protection (both $p_{\mathrm{Holm}}{<}10^{-3}$).

\subsection{Mechanism Analysis in Simulation}
\label{sec:sim}

Real data cannot show which drift a controller let through, because the true causal support of an edit is unknown. We therefore test the acceptance logic in \emph{SimEdit}, a procedural environment whose ground-truth masks are hidden from every controller (details in suppl.\ \suppref{sup:sim}{Sec.~C}).
Each $96{\times}96$ scene contains a target object with its shadow, a face-like disk, and a text-like block. The controller sees only a noisy grounding of the target (mean IoU 0.61), and a stochastic editor injects re-synthesis noise, color casts, small shifts, and identity changes.
A run is a \textbf{constrained success} (CSR) if the target score reaches 0.7 while face, text, and background each stay within an MAE of 0.01. We use 200 scenes $\times$ 3 seeds and at most 16 editor calls per run.

\begin{figure*}[t]
\centering
\includegraphics[width=\textwidth]{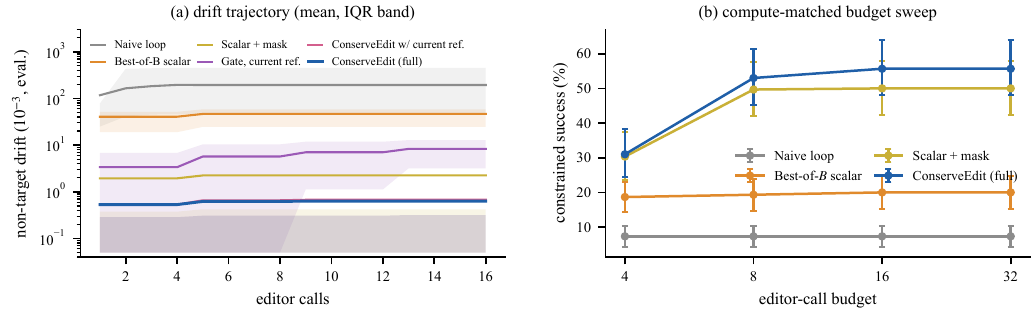}
\caption{\textbf{(a)} Evaluation drift along the trajectory (log scale): source-anchored \method stays flat, as \cref{prop:inv} predicts; without a mask, a current-image reference accumulates drift (Gate, current ref.; suppl.\ \suppref{prop:goal}{Proposition~S1}), and with a mask it coincides with \method (pink under blue). \textbf{(b)} Constrained success under matched editor-call budgets (100 held-out scenes $\times$ 3 seeds, 95\% CIs; levels differ from the 200 scenes of the text): more sampling does not help unmasked loops, and \method is at or above Scalar\,+\,mask at every budget, with overlapping CIs.}
\label{fig:traj}
\end{figure*}

\begin{figure}[t]
\centering
\includegraphics[width=\columnwidth]{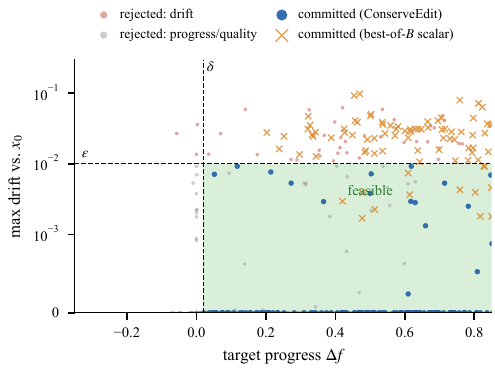}
\caption{\textbf{Acceptance geometry with measured candidates} (SimEdit, 100 scenes). Every \method commit lies in the feasible region (\cref{prop:inv}). Of the best-of-$B$ scalar loop's commits, 83.5\% exceed $\epsilon$; these are compensated trades of drift for progress (\cref{prop:scalar}).}
\label{fig:gate}
\end{figure}

\paragraph{Results.}
Accepting every output (naive loop) reaches the target but only 5.8\% CSR, and best-of-$B$ scalar selection reaches 14.5\%: 83.5\% of its commits exceed the drift tolerance (\cref{fig:gate}). Masked commitment raises CSR to 55.0\% (Scalar\,+\,mask), and the full gate to 61.8\% [56,\,67] with $4\times$ lower entity drift ($0.51$ vs.\ $2.11\times10^{-3}$). This gain comes from protected entities that the noisy support overlaps, where drift rejections trigger the shrink repair. Without a mask, a current-image reference lets drift grow with the number of calls, as \cref{sec:theory} predicts (\cref{fig:traj}a), and at every matched editor-call budget \method is at or above Scalar\,+\,mask (\cref{fig:traj}b). Grounding quality bounds all masked controllers: \method succeeds on 78.2\% of scenes with IoU $\ge0.6$ and on 41.9\% below. Full tables, budget and weight sweeps, and leakage tests are in suppl.\ \suppref{sup:sim}{Sec.~C}.

\section{Limitations and Broader Impact}
\label{sec:limits}

\paragraph{Coverage of the guarantees.} \Cref{prop:inv} holds for any executor but only with respect to the grounded support and the chosen descriptors, so grounding is the main bottleneck: in SimEdit, success falls from 78.2\% to 41.9\% below IoU 0.6, and on real images automatic masks raise abstention to 41\%, mostly on poorly grounded turns.

\paragraph{Where the gate adds value.} With a background-only contract, the gate adds nothing over masked scalar selection (same output on 92\% of turns), and one masked call matches its GT similarity. It pays off when protected entities lie near or inside a noisy support ($+6.8$ points in simulation; 53\% vs.\ 23\% constrained success on real faces and text), but that subset is small (90 turns), and the text descriptor is weak: EasyOCR changes its reading under 17\% of benign perturbations, so a string change alone never counts as a violation.

\paragraph{Predicates, seams, and abstention.} The local CLIP predicate rejects many valid edits (13\% abstention, 55\% on removals). Detector predicates calibrated on dev fix removals but abstain on 47\% of additions and lower CLIP-D to 0.054 (suppl.\ \suppref{sup:pred}{Sec.~B}); an MLLM or task-specific verifier remains the most direct fix. A seam check calibrated on GT composites rejects every committed Qwen composite: seams are systematic for a re-rendering editor and call for support adaptation or harmonization inside the halo.

\paragraph{Evaluation and scope.} GT similarity rewards conservative fills, so we report preservation, GT similarity, and instruction alignment separately; we ran no human preference study. We evaluate one editor (Qwen-Image-Edit) on one dataset (MagicBrush). The method targets local and compositional edits, and with $T{=}2$, $B{=}4$ it costs up to $8\times$ a direct call. Preservation constraints reduce unintended change but do not decide whether an edit is acceptable; the released code keeps executor safety filters.

\section{Conclusion}
\label{sec:conclusion}
\method treats each refinement as a source-anchored transaction with provable invariance and an iteration-independent drift bound.
On real edits with Qwen-Image-Edit, masked commitment alone removes global drift, and the separate tests protect faces and text that automatic grounding exposes; grounding remains the bottleneck.

\section*{LLM Usage Statement}
\paragraph{Research.} The generative models in our experiments are image editors: the real-editor study (\cref{sec:real}) runs the public, frozen Qwen-Image-Edit checkpoint~\cite{wu2025qwenimage}, which is conditioned through Qwen2.5-VL~\cite{bai2025qwen25vl}, as the object of study, and the frozen FLUX.1 Fill [dev] inpainting model~\cite{bfl2024fill} as a baseline; the EditRefiner baseline also re-edits with its own frozen Qwen-Image-Edit-2511. Their outputs appear in \cref{fig:teaser,fig:overview,fig:realqual} and in the supplementary galleries, and all of them were produced and logged by the released code. MLLMs propose instructions or verify edits only inside the published agentic baselines of \cref{sec:real}, each run with its released, frozen weights as part of that method: EditThinker-8B (a Qwen3-VL-8B~\cite{bai2025qwen3vl} fine-tune) scores each output and rewrites the prompt~\cite{li2025editthinker}; RePlan's Qwen2.5-VL-7B planner~\cite{bai2025qwen25vl} turns the instruction into region-level guidance~\cite{qu2026replan}; and EditRefiner's perception, reasoning, and evaluation agents (LoRA adapters on Qwen3-VL-8B-Instruct) localize flaws, describe them, and score each refinement~\cite{xu2026editrefiner}. \method itself uses no LLM or MLLM in any reported experiment, and no MLLM judges results: controller scores and evaluation metrics are fixed formulas on CLIP~\cite{radford2021clip}, SigLIP~\cite{zhai2023siglip}, and DINOv2~\cite{oquab2024dinov2} embeddings or on detector scores, automatic masks come from fixed parsing rules and frozen detection and segmentation models~\cite{liu2024groundingdino,kirillov2023sam}, face and text checks use frozen face-recognition and OCR models, and SimEdit (\cref{sec:sim}) is fully procedural.

\paragraph{Writing and code.} We used Claude (Anthropic) to diagnose a LaTeX layout problem in the supplementary material, to review the structure and presentation of the manuscript, to draft revised wording for parts of the abstract, introduction, related work, experiments, limitations, conclusion, and figure captions, to suggest figure and table layout changes, and to draft this statement. Following the authors' written experiment plan, an LLM coding assistant also implemented and ran the scripts for the Direct\,+\,mask baseline, the automatic-grounding study, the full-test-set run, the $1024^2$ run, the inpainting baseline, the face/text study, the predicate study, and the comparison with published agentic editors (including the compatibility changes listed in the supplementary material), assembled \cref{fig:teaser} from logged outputs, and drafted the text that reports them; every reported number is read from the logged script outputs. The authors reviewed and edited all of its output and verified every number against the experiment logs. No research ideas, results, or citations were generated by LLMs, and the authors take full responsibility for the content.

{
    \small
    \bibliographystyle{ieeenat_fullname}
    \bibliography{main}
}


\clearpage
\setcounter{page}{1}
% Supplementary pages are numbered S1, S2, ... so that hyperref's page anchors
% (and the pagebackref entries in the bibliography) do not collide with main-paper pages.
\renewcommand{\thepage}{S\arabic{page}}
\maketitlesupplementary
\renewcommand{\thesection}{\Alph{section}}
\setcounter{section}{0}
\renewcommand{\thetable}{S\arabic{table}}
\renewcommand{\thefigure}{S\arabic{figure}}
\renewcommand{\theproposition}{S\arabic{proposition}}
\setcounter{table}{0}\setcounter{figure}{0}\setcounter{proposition}{0}
% Counters are reset above, so give hyperref unique anchor names (\theH...);
% otherwise links to Sec. A-E and Prop. S1-S3 jump to Sec. 1-5 and Prop. 1-3 of the main paper.
% (\def rather than \renewcommand: hyperref may define \theHproposition only on first use.)
\def\theHsection{S\Alph{section}}
\def\theHtable{S\arabic{table}}
\def\theHfigure{S\arabic{figure}}
\def\theHproposition{S\arabic{proposition}}

This supplementary material extends the main paper. Section, figure, table, and proposition numbers without the prefix ``S'' refer to the main paper, and citation numbers refer to its reference list.
All numbers and images below come from the released code and logs of the real-editor run (\cref{sec:real}) and the SimEdit run (\cref{sec:sim}); each qualitative selection rule is stated in its caption.
Contents:
\cref{sup:impl}~implementation details and compute;
\cref{sup:real}~additional real-editor results (committed-only and stratified tables, metric distributions, abstention and false-rejection analysis, seam close-ups, 24 further qualitative turns);
\cref{sup:sim}~SimEdit details, full tables, and sweeps;
\cref{sup:theory}~complete proofs and additional guarantees;
\cref{sup:repro}~reproducibility, assets, and licenses.

%=====================================================================
\section{Implementation Details}
\label{sup:impl}

\paragraph{Contract and controller.}
\Cref{tab:s_hparams} lists every hyperparameter. The two studies share the controller code (\texttt{conserveedit/loop.py}); only the executor, predicate, and descriptors differ.
Candidates for one step share the input $x_t$ and the atomic prompt and differ in their seeds. In the real study, the per-turn seed stream is derived from $\mathrm{sha1}(\textit{img\_id}\text{-}\textit{turn})$, so all controllers draw identical first-step seeds. Identical (input, prompt, seed) calls are generated once and shared, so differences between controllers never stem from sampling luck in the first step.

\begin{table}[t]
\centering
\caption{\textbf{Hyperparameters.} Values for the controlled study (SimEdit, \cref{sec:sim}) and the real-editor study (Qwen-Image-Edit, \cref{sec:real}). All were fixed before the reported runs.}
\label{tab:s_hparams}
\scriptsize
\setlength{\tabcolsep}{3pt}
\resizebox{\columnwidth}{!}{%
\begin{tabular}{@{}lll@{}}
\toprule
Parameter & SimEdit & Real (MagicBrush) \\
\midrule
Budget $T$ / width $B$ & 4 / 4 (sweeps: $T\in\{1,2,4,8\}$) & 2 / 4 \\
Clause threshold $\tau$ / progress floor $\delta$ & 0.7 / 0.02 & 0.5 / 0.02 \\
Target predicate $f$ & redness on true target & local CLIP-B/32 directional \\
Core support $M$ & noisy grounding & human mask, or auto (GD\,+\,SAM) \\
Envelope threshold $\tau_{env}$ & -- & 0.8 (dev split) \\
Halo $H$ & 1\,px (+repairs, $\le$3\,px) & 8\,px (+repairs, $\le$3\,px) \\
Protected entities & face, text (pixel $+0.5\times$edge) & background (pixel) \\
Entity tolerance $\epsilon_k$ & 0.01 & 0.02 \\
Background tolerance $\epsilon_{bg}$ & 0.01 & 0.02 \\
Quality floor $q_{\min}$ & 0.3 & 0.3 \\
Commit-mask feathering & 2 box-blur passes, clipped to $A$ & same \\
Scalar baseline weight $\lambda$ & 1 (sweep $0.25$--$256$) & 1 \\
Stopping & clause met in 2 audits, or budget & same \\
Resolution & $96^2$ & $512^2$ \\
\bottomrule
\end{tabular}}
\end{table}

\paragraph{Real executor and evaluator.}
\emph{Executor:} Qwen/Qwen-Image-Edit at revision \texttt{ac7f931}, loaded with \texttt{QwenImageEditPipeline} (diffusers 0.42.1) in bf16; 20 inference steps, true CFG scale 4.0, negative prompt a single space, output $512^2$. The $B$ candidates of a step are generated in one batch with one generator per seed.
\emph{Controller predicate:} the image is cropped to the bounding box of the mask dilated by 8\,px, and CLIP ViT-B/32~\cite{radford2021clip} embeds the crop. The score is $f=\mathrm{clip}\big(\langle \hat d, e_{u}\rangle/0.15, 0, 1\big)$, where $\hat d$ is the normalized change of the crop embedding relative to $x_0$ and $e_u$ the instruction's text embedding.
\emph{Quality $q$:} one minus twice the mean gradient magnitude on the boundary of the commit mask, clipped to $[0,1]$.
\emph{Evaluator}, disjoint from the controller: CLIP ViT-L/14 and DINOv2-base~\cite{oquab2024dinov2}, applied to whole images and compared with the MagicBrush ground truth. CLIP-D is the cosine between the normalized whole-image ViT-L/14 embedding change and the instruction embedding.

\paragraph{Inpainting baseline.} FLUX.1 Fill [dev]~\cite{bfl2024fill} (\texttt{FluxFillPipeline}, bf16) receives the source upsampled to $1024^2$ (Lanczos), the support $A{=}M\cup H$ (nearest neighbour; white means ``may change'') and the MagicBrush instruction verbatim as its prompt; Fill expects a description of the content, and we deliberately do not rewrite the instruction. Settings follow the model card: guidance 30, 50 steps, \texttt{max\_sequence\_length} 512, one call per turn with the turn's first step-0 seed. The output is downsampled to $512^2$ and composed through the same $W$ as \method. Before composition, its region outside the mask is not preserved (PSNR$_{out}$ 35.4\,dB, from the autoencoder round trip); after composition it is bit-exact. With automatic masks, Fill uses the automatic support and returns $x_0$ when grounding abstains.

\paragraph{Published agentic editors (\cref{tab:s_agents}).}\label{sup:agents}
\emph{Selection.} We took the agentic editors discussed in \cref{sec:related} and kept those with released inference code and weights: EditThinker~\cite{li2025editthinker} (code commit \texttt{803d1a7}, weights \texttt{appletea2333/EditThinker-8B} at \texttt{1e873a5}), RePlan~\cite{qu2026replan} (\texttt{6c9b12f}, \texttt{TainU/RePlan-Qwen2.5-VL-7B} at \texttt{518f823}), and EditRefiner~\cite{xu2026editrefiner} (\texttt{b17cea6}, \texttt{TmpAccount/EditRefiner} at \texttt{6afa121}). MIRA's repository contains only a README, and Agent Banana's announces its code as forthcoming, so neither could be run.
\emph{Common protocol.} All three run on the first 160 turns of the fixed permutation (the subset of \cref{tab:s_res}), at $512^2$, with the turn's seed sequence of \method and at most eight editor calls (${=}\,T B$). Each method gives two rows: its raw output, and that output composed through the same $W$ as Direct\,+\,mask (human mask plus 8\,px halo, feathered), which isolates the effect of pasting back. Calls in \cref{tab:s_agents} count editor calls; MLLM calls are reported below.
\emph{EditThinker} runs its official think-while-edit loop (prompt template and stopping rule of \texttt{run\_edit\_thinker\_inference.py}, \texttt{-{}-skip\_first\_turn\_rewrite}): turn 1 edits $x_0$ with the original instruction using our Qwen-Image-Edit (\cref{sup:impl} settings; turn $k$ uses the $k$-th seed, so turn 1 is the Direct call); EditThinker-8B then scores the last output against $x_0$ and rewrites the prompt, and the next turn re-edits $x_0$. It stops when the prompt is unchanged or both scores reach 8; the output is the last edit. We raise the default limit of 5 turns to 8 to match the budget. The thinker runs with transformers instead of vLLM, using its generation config (as vLLM does by default) with a fixed per-call seed and at most 4{,}096 new tokens.
\emph{RePlan} runs the official \texttt{RePlanPipeline} with \texttt{pipeline\_type} ``qwen'' (Qwen-Image-Edit weights with region-attention injection) and the recommended attention switch 0.5: one planner call, then one edit with 20 steps and true CFG 4.0, forced to $512^2$. Its pipeline code requires the older diffusers 0.37.1, which we load for this method only, and the planner uses SDPA instead of FlashAttention-2 (not installed; the same attention function).
\emph{EditRefiner} refines a given edit, so we hand it the Direct output (one call) and run the official \texttt{run\_pipeline} of \texttt{inference\_main.py} with \texttt{max\_iters} 7 (at most $1{+}7$ editor calls). Each iteration runs the perception agent (two saliency models on the initial edit), the reasoning agent (flaw descriptions), a Qwen-Image-Edit-2511 re-edit of the current image (source, current image, and flaw mask as inputs; 40 steps, seed 0, the method's own editor), and the evaluation agent. An iteration is kept only if its score improves, and the loop stops at the first non-improvement. Three changes were needed to run the released code in our environment (transformers 5.22.1 instead of the pinned 5.3.0), none of which alters a model or a decision rule: the reasoning agent's \texttt{swift infer} call is executed with transformers and peft (same adapter, chat template, and greedy decoding as its training arguments); Qwen3-VL's multimodal token-type ids, which transformers 5.22 requires and the agents' code does not pass, are rebuilt from the input ids exactly as the processor would; and the evaluation agent receives the local path of the same Qwen3-VL-8B-Instruct base model, because its path resolver otherwise fails. The 2511 pipeline is loaded once per worker instead of once per turn, and diffusers 0.37.1 (the method's pinned version) is used.
\emph{Calls.} EditThinker uses 1 to 8 editor calls (26 turns stop after the first edit, 53 after two, and 25 reach the cap of 8), on average 3.55 editor and 3.39 thinker calls. RePlan uses one planner and one editor call. EditRefiner uses 1 to 4 refinement calls (119 turns stop after one, 26 after two, 13 after three, 2 after four), on average 2.36 editor calls including the Direct call and 6.45 MLLM calls (perception counted as two), and its output differs from the Direct output on 41 turns; with the agents reloaded in every iteration, as released, a turn takes 285\,s on average on one H200.

\begin{table}[t]
\centering
\caption{\textbf{Published agentic editors} (Qwen-Image-Edit, the first 160 turns of the fixed permutation; moved from \cref{tab:real}). ``+\,same $W$'' composes the method's output through the commit mask of \method. The last two rows are the main-paper controllers on the same turns. Calls: editor calls (MLLM calls in the text).}
\label{tab:s_agents}
\scriptsize
\setlength{\tabcolsep}{2.4pt}
\resizebox{\columnwidth}{!}{% Moved from the main Tab. 1 (bottom group). Generated by code/experiments/make_table_agents.py from results_v2/agents/table_agents_rows.tex (first 160 turns of the fixed permutation).
\begin{tabular}{@{}lccccccc@{}}
\toprule
Controller & CLIP-D$\uparrow$ & Abstain & L1$_{gt}\!\downarrow$ & CLIP-I$_{gt}\!\uparrow$ & DINO$_{gt}\!\uparrow$ & PSNR$_{out}\!\uparrow$ & Calls \\
\midrule
EditThinker~\cite{li2025editthinker} & 0.062 & 0\% & 0.184 & 0.833 & 0.708 & 12.5 & 3.5 \\
\quad + same $W$ & 0.069 & 0\% & 0.042 & 0.900 & 0.871 & exact & 3.5 \\
RePlan~\cite{qu2026replan} & 0.044 & 0\% & 0.080 & 0.901 & 0.837 & 23.2 & 1.0 \\
\quad + same $W$ & 0.044 & 0\% & 0.035 & 0.931 & 0.882 & exact & 1.0 \\
EditRefiner~\cite{xu2026editrefiner} & 0.061 & 0\% & 0.182 & 0.828 & 0.697 & 13.0 & 2.4 \\
\quad + same $W$ & 0.070 & 0\% & 0.042 & 0.904 & 0.869 & exact & 2.4 \\
\midrule
Direct\,+\,mask (same turns) & 0.070 & 0\% & 0.042 & 0.898 & 0.865 & exact & 1.0 \\
{\method} (same turns) & 0.079 & 18\% & 0.041 & 0.898 & 0.867 & exact & 6.8 \\
\bottomrule
\end{tabular}
}
\end{table}

\paragraph{Automatic grounding.}\label{sup:ground}
\emph{Phrase:} spaCy \texttt{en\_core\_web\_sm} with fixed rules, no LLM. An instruction is an \emph{add} edit if it starts with add, put, place, or insert, or with ``let/have/make there be''. Otherwise the target is the first noun chunk whose head is neither a pronoun nor a generic word (image, photo, color, material, side, \ldots), with determiners removed. For add edits the region is the box of the anchor noun after a spatial preposition (on, in, near, next to, under, \ldots), enlarged by 25\% per side; without an anchor the turn abstains.
\emph{Masks:} Grounding DINO-base~\cite{liu2024groundingdino} (box threshold 0.35, text threshold 0.25) keeps the highest-scoring box. SAM ViT-H~\cite{kirillov2023sam} segments it and four copies shifted uniformly by up to $\pm5\%$ of the box width and height (per-turn seed $\mathrm{sha1}(\textit{img\_id}\text{-}\textit{turn})$). The envelope is the minimum IoU between a shifted mask and the main mask; add edits use the box region and skip this test. An abstaining turn returns $x_0$ for every masked controller.
\emph{Calibration on dev} (528 turns; no test data): 58 turns have no phrase, 70 no detection, 70 are add boxes, and 330 reach the envelope test. The envelope rejects 3.6, 5.2, 7.3, and 8.8\% of these at $\tau_{env}{=}0.5$, 0.6, 0.7, and 0.8. The rule fixed beforehand, the largest $\tau_{env}$ with at most 20\% rejections, gives $\tau_{env}{=}0.8$. Mean IoU with the human mask over the kept dev turns is 0.30 for every $\tau_{env}$ (0.33 for SAM masks alone). In a manual check of the first 20 dev parses, 18 name the intended target; the two errors pick ``place'' in ``put a lion in the place of the donkey'' and ``drawing'' in ``a drawing of a bird on his shirt''.

\paragraph{Face and text entities (\cref{tab:entities}).}\label{sup:entities}
\emph{Subset}, fixed before any run (rule and metrics in \texttt{preregistration.json}): on every test source we detect faces with insightface \texttt{buffalo\_l} (SCRFD, score $\ge$0.6, side $\ge$16\,px) and text with EasyOCR (confidence $\ge$0.5, $\ge$2 alphanumeric characters). An instance is protected if its box does not intersect the human mask $M$ and intersects, or lies within 32\,px of, the human support or the automatic support. Of the 1{,}053 turns, 293 contain a face or text instance and 90 have at least one protected instance (93 faces, 28 text instances).
\emph{Descriptors} (one protected entity per instance, with its box as mask): face, $D{=}1{-}\cos$ of ArcFace R50 (WebFace600K) embeddings of $x_0$ and $x$, both aligned with the source's five keypoints; text, $D{=}[\text{EasyOCR string of }x\ne\text{that of }x_0]+\text{pixel MAE in the box}$. Each descriptor returns its distance to $x_0$ scaled by the entity's pixel count, so the unchanged audit of \cref{eq:drift} yields $D$ exactly; Scalar\,+\,mask adds the same terms to its drift ($\lambda{=}1$), and the shrink repair removes a violated entity from the support.
\emph{Tolerances}, calibrated on the 528 dev sources (195 faces, 82 text instances) under JPEG $q{=}90$, brightness $\pm2\%$, and 1\,px shifts as the 99th percentile of $D$: $\epsilon_{face}{=}0.111$ and $\epsilon_{text}{=}1.089$. EasyOCR changes its reading under 17\% of these benign perturbations, so a string change alone never exceeds $\epsilon_{text}$.
\emph{Evaluator}, disjoint from the controller: ArcFace R100 (Glint360K, insightface \texttt{antelopev2}) with $1{-}\cos{>}0.150$, and TrOCR-base-printed~\cite{li2023trocr} with $1{-}$string similarity${>}0.333$, both thresholds calibrated in the same way. The insightface ONNX models ran on the CPU execution provider throughout (calibration, selection, runs, evaluation).
\emph{Details}: with automatic masks, Direct\,+\,mask violates 41\% of the protected faces and 46\% of the text instances, \method 2\% and 0\%; \method's shrink repair fires on 44 of the 90 turns.

\paragraph{Repairs.} After a full rejection, the controller applies the first applicable rule:
(i)~if a protected entity was violated and overlaps the support, remove it from the support (\emph{shrink});
(ii)~else, if every candidate failed only the progress floor and fewer than three expansions have been used, dilate the support by 1\,px while excluding every protected entity (\emph{halo expansion}, logged as a contract revision);
(iii)~else, resample seeds.
An action signature (clause, prompt, seed) that was already tried is never retried.

\paragraph{Compute.}
The real study ran on H200 GPUs with two workers per GPU. Running all six controllers on one turn took a median of 354\,s; the 1{,}053 turns used 105 worker-hours. The shared cache cut the cost to 14{,}973 unique Qwen-Image-Edit calls, or 14.2 per turn, instead of 33.0 calls per turn without sharing.
Mean editor calls per controller were: direct 1.0, naive loop 3.4, best-of-$B$ scalar 6.9, scalar\,+\,mask 6.8, gate without mask 8.0, and \method 6.9.
Direct\,+\,mask in \cref{tab:real} reuses the cached direct outputs and needs no editor call.
The human-mask counterparts of \cref{tab:auto} are the rows of \cref{tab:real}. Through a persistent call cache, the auto-mask rows reuse their step-0 candidates and needed only 2{,}100 new calls (13.2 worker-hours). The face and text study reused cached candidates and needed 104 new editor calls (2.2 worker-hours); the predicate study (three variants, 160 and 1{,}053 turns) needed 836 (8.3 worker-hours). FLUX.1 Fill needed one call per turn and mask source (median 18\,s with two workers per GPU; 4.6 worker-hours for the human masks). The $1024^2$ run of \cref{tab:s_res} (five controllers, 160 turns, one worker per GPU) needed 1{,}447 calls and 7.4 GPU-hours, a median of 201\,s per turn.
The complete SimEdit study (main table, all sweeps, figures) runs on a laptop CPU in 170\,s.

%=====================================================================
\section{Additional Real-Editor Results}
\label{sup:real}

\paragraph{Committed turns only.}
\Cref{tab:real} counts every abstention as an unedited output. \Cref{tab:s_committed} restricts all controllers to the 917 turns on which \method committed. On these turns, masked commitment raises the evaluator's CLIP-D from 0.066 (direct) to 0.094 and DINO similarity to the GT from 0.660 to 0.850, with zero change outside the support. The scalar\,+\,mask loop also commits on all 917 turns, and the two masked controllers are again nearly identical. Direct\,+\,mask, a single direct call composed through the same commit mask, reaches the same GT similarity (DINO$_{gt}$ 0.850, L1$_{gt}$ 0.048) at a lower CLIP-D (0.078); paired tests are in \cref{sec:real}.

\begin{table}[t]
\centering
\caption{\textbf{Real editor, restricted to the 917 turns where \method commits.} Same metrics as \cref{tab:real}. The ``Gate, no mask'' row never commits, so its GT similarities are those of the unedited source.}
\label{tab:s_committed}
\scriptsize
\setlength{\tabcolsep}{2.5pt}
\resizebox{\columnwidth}{!}{\begin{tabular}{@{}lcccccc@{}}
\toprule
Controller & CLIP-D$\uparrow$ & Abstain & CLIP-I$_{gt}\!\uparrow$ & DINO$_{gt}\!\uparrow$ & L1$_{gt}\!\downarrow$ & PSNR$_{out}\!\uparrow$ \\
\midrule
Direct (1 call) & 0.066 & 0\% & 0.817 & 0.660 & 0.198 & 12.1 \\
Naive loop & 0.066 & 0\% & 0.734 & 0.489 & 0.262 & 9.9 \\
Best-of-$B$ scalar & 0.065 & 17\% & 0.839 & 0.703 & 0.165 & 20.5 \\
Direct\,+\,mask (1 call) & 0.078 & 0\% & 0.893 & 0.850 & 0.048 & exact \\
Scalar\,+\,mask & 0.094 & 0\% & 0.889 & 0.850 & 0.049 & exact \\
Gate, no mask & 0.000 & 100\% & 0.930 & 0.870 & 0.036 & exact \\
\method & 0.094 & 0\% & 0.890 & 0.850 & 0.049 & exact \\
\bottomrule
\end{tabular}
}
\end{table}

\paragraph{Automatic grounding in detail.}
In \cref{tab:auto}, 317 of the 1{,}053 instructions parse as add edits and 736 as other edits. Grounding succeeds on 710 turns (541 SAM masks, 169 add boxes); 140 have no phrase, 161 no detection, and 42 fail the envelope test. Mean IoU with the human mask is 0.35 for SAM masks and 0.18 for add boxes, and the automatic masks are larger (24\% vs.\ 15\% of the image). On all 710 grounded turns, the three masked controllers leave every pixel outside their own support (including halo repairs) bit-identical to the source. After grounding, \method abstains on another 89 turns (8.5\% of 1{,}053). Split by IoU, automatic and human masks give the same \method result on the 139 turns with IoU${\ge}0.6$ (CLIP-D 0.062 vs.\ 0.061, DINO$_{gt}$ 0.809 vs.\ 0.807, all $p_{\mathrm{Holm}}{\ge}0.5$). On the 571 turns below 0.6 they differ (DINO$_{gt}$ 0.777 vs.\ 0.866, L1$_{gt}$ 0.075 vs.\ 0.041, $p_{\mathrm{Holm}}{<}10^{-3}$).

\paragraph{Hardware.} Editor calls are reproduced bit-exactly on the same GPU type: re-running three turns of the original run on a server with the same GPUs (H200; driver 610.57 instead of 580.178) gives pixel-identical outputs for all six controllers. The 893 turns added to the original 160 and all automatic-mask runs in \cref{tab:auto} were therefore produced on this GPU type. On a different GPU type (H200 NVL), candidates differ numerically (mean $|\Delta|$ of 1.6--3.8 grey levels for the direct output of the same three turns). When the original 160 turns were re-run there with human masks, commit decisions nonetheless agreed on 100\%, 98.8\%, and 96.9\% of turns for Direct\,+\,mask, scalar\,+\,mask, and \method, and no GT or CLIP metric differed after Holm correction (\method: CLIP-D 0.080 vs.\ 0.079, DINO$_{gt}$ 0.867 vs.\ 0.867).

\paragraph{Working resolution.}
MagicBrush sources and targets are $512^2$, whereas current editors work best at about one megapixel. \Cref{tab:s_res} repeats the study at $1024^2$ on the original 160-turn subset: the source is upsampled bicubically, the mask with nearest neighbour, the halo and the predicate box scale to 16\,px, and composition happens at $1024^2$. Outputs are then downsampled (Lanczos) and compared with the original $512^2$ GT, and an abstention at $1024^2$ counts as $x_0$. This tests whether drift and conclusions change when the editor runs at its preferred resolution; it adds no real detail. Direct editing drifts much less at $1024^2$ (PSNR$_{out}$ 22.3 vs.\ 12.2\,dB, DINO$_{gt}$ 0.788 vs.\ 0.678), and the naive loop still compounds drift (19.3\,dB). The masked controllers remain bit-exact outside the support at $1024^2$; at $512^2$ their PSNR$_{out}$ (48--50\,dB) reflects the resampling round trip, whose floor is 49.2\,dB. \method still beats direct editing in every GT metric (DINO$_{gt}$ 0.880 vs.\ 0.788, L1$_{gt}$ 0.037 vs.\ 0.094, $p_{\mathrm{Holm}}{<}10^{-3}$), although its CLIP-D advantage over direct editing is no longer significant (0.077 vs.\ 0.073, $p{=}0.43$). As at $512^2$, it matches Direct\,+\,mask in GT similarity (DINO$_{gt}$ 0.880 vs.\ 0.887, $p_{\mathrm{Holm}}{=}0.58$) and exceeds it in CLIP-D (0.077 vs.\ 0.064, $p_{\mathrm{Holm}}{=}0.002$), and it abstains on 16\% of turns (18\% at $512^2$).

\begin{table}[t]
\centering
\caption{\textbf{Working resolution} (Qwen-Image-Edit, the 160-turn subset; $512^2$ rows are the original run). At $1024^2$ the editor, the halo (16\,px), and composition run at $1024^2$; GT metrics are computed after Lanczos downsampling to $512^2$. PSNR$_{out}^{512}$ of a bit-exact $1024^2$ output is bounded by the resampling floor, 49.2\,dB.}
\label{tab:s_res}
\scriptsize
\setlength{\tabcolsep}{2.2pt}
\resizebox{\columnwidth}{!}{% Generated by code/experiments/res_eval.py from results_v2/qwen1024 and results_v2/qwen_human (n=160).
\begin{tabular}{@{}llcccccc@{}}
\toprule
Res. & Controller & CLIP-D$\uparrow$ & Abstain & DINO$_{gt}\!\uparrow$ & L1$_{gt}\!\downarrow$ & PSNR$_{out}^{512}$ & PSNR$_{out}^{1024}$ \\
\midrule
$512^2$ & Direct (1 call) & 0.061 & 0\% & 0.678 & 0.193 & 12.2 & -- \\
$512^2$ & Naive loop & 0.059 & 0\% & 0.477 & 0.258 & 9.9 & -- \\
$512^2$ & Direct\,+\,mask (1 call) & 0.070 & 0\% & 0.865 & 0.042 & exact & -- \\
$512^2$ & Scalar\,+\,mask & 0.081 & 15\% & 0.865 & 0.041 & exact & -- \\
$512^2$ & \method & 0.079 & 18\% & 0.867 & 0.041 & exact & -- \\
\midrule
$1024^2$ & Direct (1 call) & 0.073 & 0\% & 0.788 & 0.094 & 22.3 & 22.3 \\
$1024^2$ & Naive loop & 0.070 & 0\% & 0.750 & 0.124 & 19.3 & 19.2 \\
$1024^2$ & Direct\,+\,mask (1 call) & 0.064 & 0\% & 0.887 & 0.037 & 48.0 & exact \\
$1024^2$ & Scalar\,+\,mask & 0.077 & 15\% & 0.881 & 0.037 & 49.6 & exact \\
$1024^2$ & \method & 0.077 & 16\% & 0.880 & 0.037 & 49.7 & exact \\
\bottomrule
\end{tabular}
}
\end{table}

\paragraph{Where \method and scalar\,+\,mask differ.}
The two masked controllers differ on 80 of 1{,}053 turns. On 12 turns \method abstains while the scalar loop commits: the scalar loop accepts any positive $J$-improvement, whereas the gate requires $\Delta f\ge\delta$. On 68 turns both commit but pick different candidates. Over the 80 turns, \method has the higher DINO similarity to the GT on 48 and the higher CLIP-D on 26. Over all turns the differences are small (\cref{sec:real}).

\paragraph{Stratified results.}
\Cref{tab:s_strata} splits the real study by mask area (terciles at 6.4\% and 16.5\% of the image) and by a keyword heuristic over the instruction text. The heuristic never looks at outputs; its rules (\texttt{edit\_type} in \texttt{experiments/suppl\_real.py}) are applied in order: remove, add, attribute, replace, pose/action, with add as the default. Two patterns stand out.
(i)~The benefit of masked commitment in GT similarity is largest for small masks (DINO$_{gt}$ 0.696$\to$0.935), where re-rendering the rest of the image dominates the error.
(ii)~Removal edits are the weakest category. \method abstains on 55\% of them, and even the direct edit has negative CLIP-D. A directional CLIP score toward the text ``remove the X'' is a poor progress signal, because the text embedding still contains the object to be removed. Removal therefore needs a different predicate, \eg, open-vocabulary detection of the object's absence.

\begin{table}[t]
\centering
\caption{\textbf{Real editor, stratified.} Mask-area terciles and edit types. $^\ast$Edit type is a keyword heuristic over the instruction text (add 590, replace 145, attribute 135, pose/action 108, remove 75).}
\label{tab:s_strata}
\scriptsize
\setlength{\tabcolsep}{2.2pt}
\resizebox{\columnwidth}{!}{\begin{tabular}{@{}llccccc@{}}
\toprule
Stratum & $n$ & Controller & CLIP-D$\uparrow$ & Abstain & DINO$_{gt}\!\uparrow$ & L1$_{gt}\!\downarrow$ \\
\midrule
Mask area: small & 351 & Direct (1 call) & 0.066 & 0\% & 0.696 & 0.186 \\
 &  & Direct\,+\,mask (1 call) & 0.065 & 0\% & 0.931 & 0.014 \\
 &  & Scalar\,+\,mask & 0.082 & 15\% & 0.933 & 0.014 \\
 &  & \method & 0.080 & 16\% & 0.935 & 0.014 \\
\addlinespace[1pt]
Mask area: medium & 351 & Direct (1 call) & 0.059 & 0\% & 0.684 & 0.187 \\
 &  & Direct\,+\,mask (1 call) & 0.073 & 0\% & 0.874 & 0.033 \\
 &  & Scalar\,+\,mask & 0.087 & 10\% & 0.873 & 0.034 \\
 &  & \method & 0.087 & 12\% & 0.874 & 0.033 \\
\addlinespace[1pt]
Mask area: large & 351 & Direct (1 call) & 0.047 & 0\% & 0.618 & 0.216 \\
 &  & Direct\,+\,mask (1 call) & 0.068 & 0\% & 0.754 & 0.093 \\
 &  & Scalar\,+\,mask & 0.079 & 10\% & 0.741 & 0.095 \\
 &  & \method & 0.078 & 11\% & 0.742 & 0.094 \\
\midrule
Edit type$^\ast$: add & 590 & Direct (1 call) & 0.075 & 0\% & 0.653 & 0.199 \\
 &  & Direct\,+\,mask (1 call) & 0.081 & 0\% & 0.854 & 0.046 \\
 &  & Scalar\,+\,mask & 0.096 & 7\% & 0.850 & 0.047 \\
 &  & \method & 0.094 & 8\% & 0.851 & 0.047 \\
\addlinespace[1pt]
Edit type$^\ast$: replace & 145 & Direct (1 call) & 0.023 & 0\% & 0.605 & 0.201 \\
 &  & Direct\,+\,mask (1 call) & 0.043 & 0\% & 0.792 & 0.061 \\
 &  & Scalar\,+\,mask & 0.053 & 18\% & 0.778 & 0.060 \\
 &  & \method & 0.052 & 19\% & 0.779 & 0.059 \\
\addlinespace[1pt]
Edit type$^\ast$: attribute & 135 & Direct (1 call) & 0.059 & 0\% & 0.747 & 0.192 \\
 &  & Direct\,+\,mask (1 call) & 0.084 & 0\% & 0.867 & 0.052 \\
 &  & Scalar\,+\,mask & 0.094 & 7\% & 0.873 & 0.052 \\
 &  & \method & 0.094 & 7\% & 0.874 & 0.052 \\
\addlinespace[1pt]
Edit type$^\ast$: pose/action & 108 & Direct (1 call) & 0.070 & 0\% & 0.661 & 0.209 \\
 &  & Direct\,+\,mask (1 call) & 0.069 & 0\% & 0.894 & 0.032 \\
 &  & Scalar\,+\,mask & 0.083 & 7\% & 0.901 & 0.032 \\
 &  & \method & 0.081 & 8\% & 0.901 & 0.032 \\
\addlinespace[1pt]
Edit type$^\ast$: remove & 75 & Direct (1 call) & -0.032 & 0\% & 0.750 & 0.151 \\
 &  & Direct\,+\,mask (1 call) & -0.005 & 0\% & 0.886 & 0.037 \\
 &  & Scalar\,+\,mask & 0.018 & 55\% & 0.862 & 0.039 \\
 &  & \method & 0.016 & 55\% & 0.864 & 0.039 \\
\bottomrule
\end{tabular}
}
\end{table}

\begin{figure*}[t]
\centering
\includegraphics[width=\textwidth]{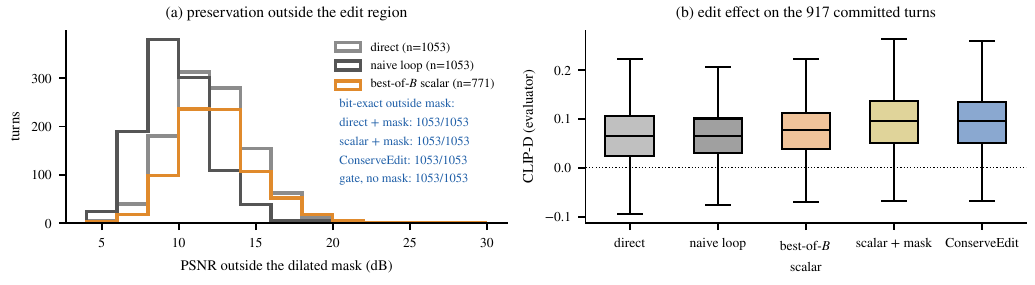}
\caption{\textbf{Per-turn metric distributions (real editor, 1{,}053 turns).} (a)~PSNR outside the dilated mask for the unmasked controllers; best-of-$B$ scalar is shown for the 771 turns on which it committed. The masked controllers and the unmasked gate are bit-exact outside the mask on all 1{,}053 turns. (b)~Evaluator CLIP-D on committed turns (box: quartiles; whiskers: 1.5 IQR). The unmasked gate never commits and is omitted.}
\label{fig:s_dist}
\end{figure*}

\paragraph{Per-turn distributions.}
\Cref{fig:s_dist} shows the per-turn values behind the means of \cref{tab:real}. Every unmasked output changes the region outside the mask (PSNR below 28\,dB on every turn), whereas the masked controllers are bit-exact there on all 1{,}053 turns. On committed turns, all three CLIP-D quartiles of the two masked controllers lie above those of direct editing.

\begin{figure*}[t]
\centering
\includegraphics[width=\textwidth]{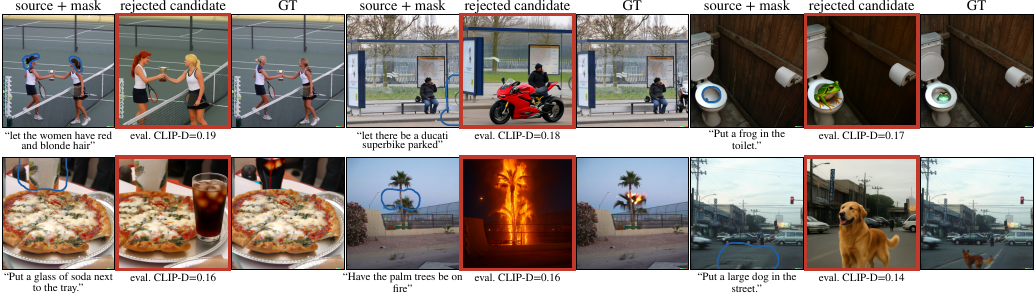}
\caption{\textbf{False rejections behind abstention.} Rule: among the 136 turns on which \method abstained, the six whose rejected first candidate (red frame; identical to the direct output) has the highest evaluator CLIP-D. Each candidate fulfils the instruction, but the local CLIP-B/32 progress predicate scored it below $\delta$. A stronger verifier (\eg, an MLLM judge or open-vocabulary detection) is the most direct fix.}
\label{fig:s_abstain}
\end{figure*}

\paragraph{Abstention and false rejections.}
\method abstains on 136 of 1{,}053 turns (12.9\%). \Cref{fig:s_abstain} shows the six abstained turns whose first rejected candidate scores highest on the evaluator's CLIP-D. That candidate is exactly the direct output, because both use the same seed and the call is shared. All six candidates are plausible edits, among them red and blonde hair, a parked Ducati, a frog in a toilet, and palm trees on fire. The local CLIP predicate nevertheless scored their progress below $\delta$. These are \emph{false rejections} of the controller's progress test, not of the preservation test, and they are the main source of abstention.

\begin{figure}[t]
\centering
\includegraphics[width=\columnwidth]{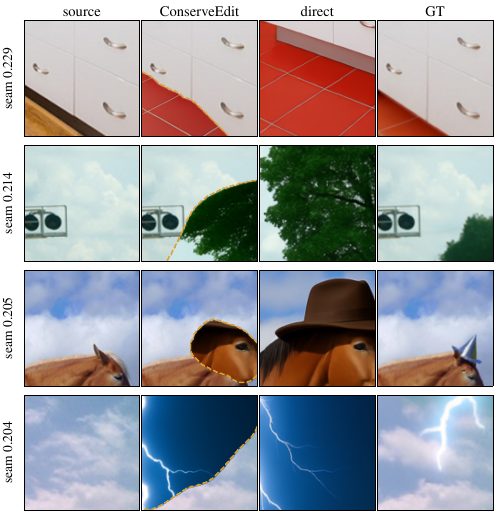}
\caption{\textbf{Worst seams.} Rule: the four committed turns with the largest seam score (left). $128^2$ crops at the strongest boundary excess; the dashed line marks the commit support. The direct output shows what the editor intended, and \method's composite truncates it at the support boundary.}
\label{fig:s_seams}
\end{figure}

\paragraph{Seams.}
We define a seam score as the mean positive excess of the image-gradient magnitude of the \method output over that of the source, on the boundary ring of the commit support. Its median over the 917 committed turns is 0.073. \Cref{fig:s_seams} shows the four highest-scoring turns. In each case, the direct output shows the editor's change reaching beyond the support, so masked commitment truncates it: a tiled floor ends at the support boundary, a tree is pasted into the sky, a hat is clipped, and a lightning-lit sky is cut off. The boundary-gradient quality check ($q_{\min}{=}0.3$) passed all four, so it is too weak for real composites. Support adaptation (moving the support with the object) or harmonization inside the halo is needed.

\begin{figure*}[t]
\centering
\includegraphics[width=\textwidth]{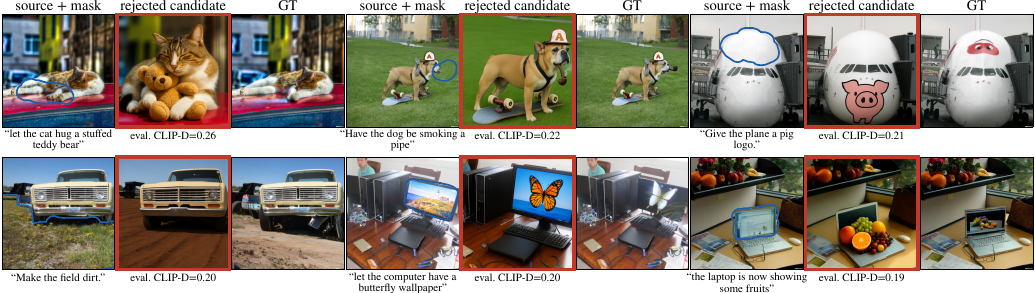}
\caption{\textbf{False rejections with the detector predicate} (corrected $\tau$). Rule as in \cref{fig:s_abstain}: among the 311 turns on which this variant abstained, the six whose rejected first candidate has the highest evaluator CLIP-D.}
\label{fig:s_abstain_det}
\end{figure*}

\begin{figure*}[t]
\centering
\begin{minipage}{\columnwidth}
\centering
\includegraphics[width=\columnwidth]{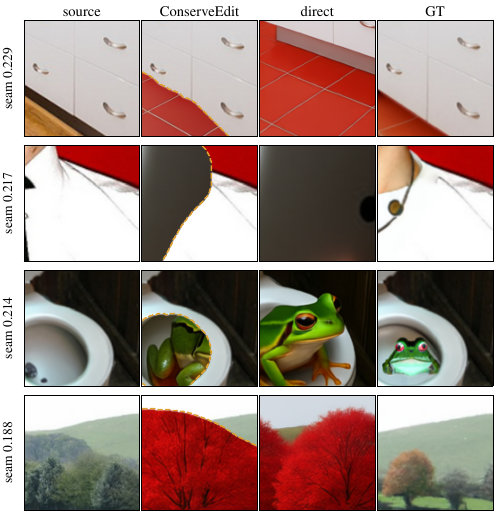}
\caption{\textbf{Worst seams with the detector predicate} (corrected $\tau$). Rule as in \cref{fig:s_seams}.}
\label{fig:s_seams_det}
\end{minipage}
\end{figure*}

\paragraph{Stronger progress predicates and seam check.}\label{sup:pred}
\emph{Predicates} (no LLM; edit type from the keyword rule above, object phrases from fixed spaCy rules): remove, the relative drop of the Grounding DINO score of the object phrase inside a crop around the support; add, the Grounding DINO score of the added object's phrase in that crop; replace, the new object's score minus the old one's; attribute and pose, the paper's local directional score with SigLIP so400m~\cite{zhai2023siglip} instead of CLIP-B/32 (CLIP-L/14 is the evaluator). A missing phrase falls back to SigLIP. \emph{Seam check}: the seam score defined below must not exceed $s_{\max}$; a failure triggers the halo-expansion repair.
\emph{Calibration on the 528 dev turns}, with each GT target composed through $W$ as a candidate; the rule fixed beforehand sets $\delta$ to the 5th percentile of its progress (floored at 0), $\tau$ to the 10th percentile of its score, and $s_{\max}$ to the 95th percentile of its seam score. This gives $\delta{=}0$ for every family, $\tau{=}0.35$ for additions but $\tau{=}0$ for the other three families (at least 10\% of correct edits score 0), and $s_{\max}{=}0.0032$ (GT composites: median 0.0009). With $\tau{=}0$ the clause holds at $x_0$, so the controller stops without editing; we therefore also report a correction computed on dev only, $\tau$ at the 10th percentile of the GT edits the predicate registers (score above that of $x_0$): 0.39 (add), 0.11 (remove), 0.02 (replace), and 0.12 (SigLIP). For reference, the paper's CLIP-B/32 predicate passes only 66\% of the dev GT edits at $\delta{=}0.02$, and 16\% reach $\tau{=}0.5$.
\emph{Results} (\cref{tab:s_pred}): with the rule as fixed, \method abstains on 76\% of turns, and on 99.9\% once the seam check is added; every committed Qwen composite of \cref{tab:real} exceeds $s_{\max}$. With the corrected $\tau$, \method abstains on 30\% of turns: never on removal, attribute, or pose edits, but on 47\% of additions; 73\% of the 136 turns that abstained in \cref{tab:real} now commit. \Cref{fig:s_abstain_det,fig:s_seams_det} repeat the selection rules of \cref{fig:s_abstain,fig:s_seams} for this variant: its false rejections are now mostly additions whose object the detector does not score above $\tau$, and the worst seams are unchanged in kind.

\begin{table}[t]
\centering
\caption{\textbf{Progress predicates and seam check} (Qwen-Image-Edit, 1{,}053 MagicBrush test turns, human masks). Abstention overall and by edit type (\%). S+M: Scalar\,+\,mask with the same predicate. ``rule as fixed'': thresholds from the calibration rule set before running; ``corrected $\tau$'': $\tau$ recomputed on dev over the GT edits the predicate registers.}
\label{tab:s_pred}
\scriptsize
\setlength{\tabcolsep}{2pt}
\resizebox{\columnwidth}{!}{% Generated by code/experiments/make_table_pred.py from results_v2/pred_full/analysis_pred.json (n=1053).
\begin{tabular}{@{}llcccccccc@{}}
\toprule
Predicate & Ctrl. & Abstain & add & repl. & attr. & pose & remove & CLIP-D$\uparrow$ & DINO$_{gt}\!\uparrow$ \\
\midrule
CLIP-B/32 (paper) & S+M & 12\% & 7 & 18 & 7 & 7 & 55 & 0.083 & 0.849 \\
 & \method & 13\% & 8 & 19 & 7 & 8 & 55 & 0.081 & 0.850 \\
\addlinespace[1pt]
detector, rule as fixed & S+M & 76\% & 57 & 100 & 100 & 100 & 100 & 0.026 & 0.858 \\
 & \method & 76\% & 57 & 100 & 100 & 100 & 100 & 0.026 & 0.859 \\
\addlinespace[1pt]
detector + seam, rule as fixed & S+M & 76\% & 57 & 100 & 100 & 100 & 100 & 0.026 & 0.858 \\
 & \method & 100\% & 100 & 100 & 100 & 100 & 100 & 0.000 & 0.868 \\
\addlinespace[1pt]
detector, corrected $\tau$ & S+M & 32\% & 47 & 32 & 4 & 6 & 9 & 0.055 & 0.858 \\
 & \method & 30\% & 47 & 23 & 0 & 0 & 0 & 0.054 & 0.858 \\
\bottomrule
\end{tabular}
}
\end{table}

\begin{figure*}[p]
\centering
\includegraphics[width=\textwidth,height=0.9\textheight,keepaspectratio]{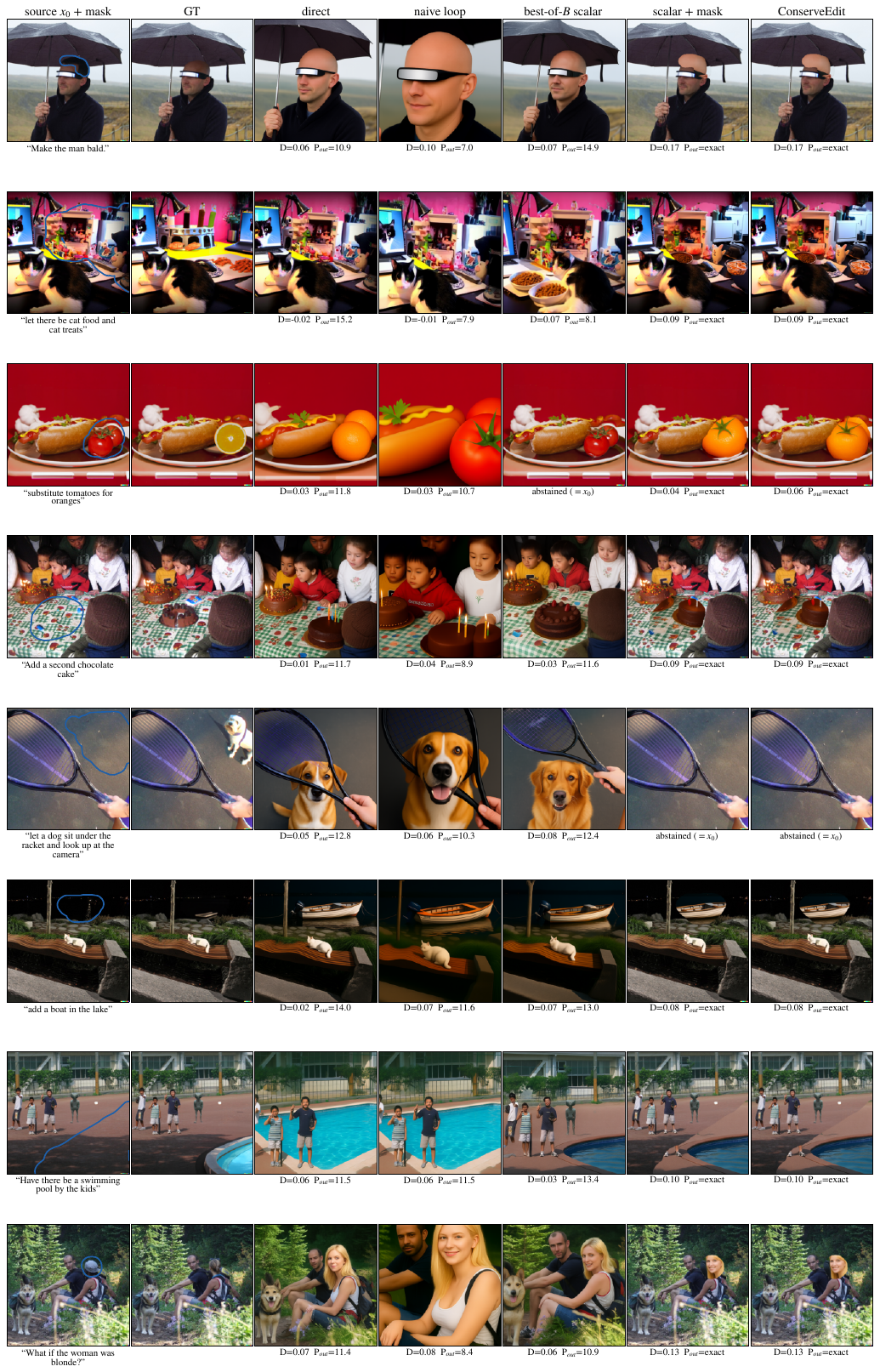}
\caption{\textbf{Additional real Qwen-Image-Edit results, part 1 of 3} (turns 1--8 of the fixed sha1-ordered list; outputs not curated). Captions give the evaluator CLIP-D (D) and PSNR outside the dilated mask (P$_{out}$; ``exact'' means bit-identical to $x_0$).}
\label{fig:s_grid1}
\end{figure*}

\begin{figure*}[p]
\centering
\includegraphics[width=\textwidth,height=0.9\textheight,keepaspectratio]{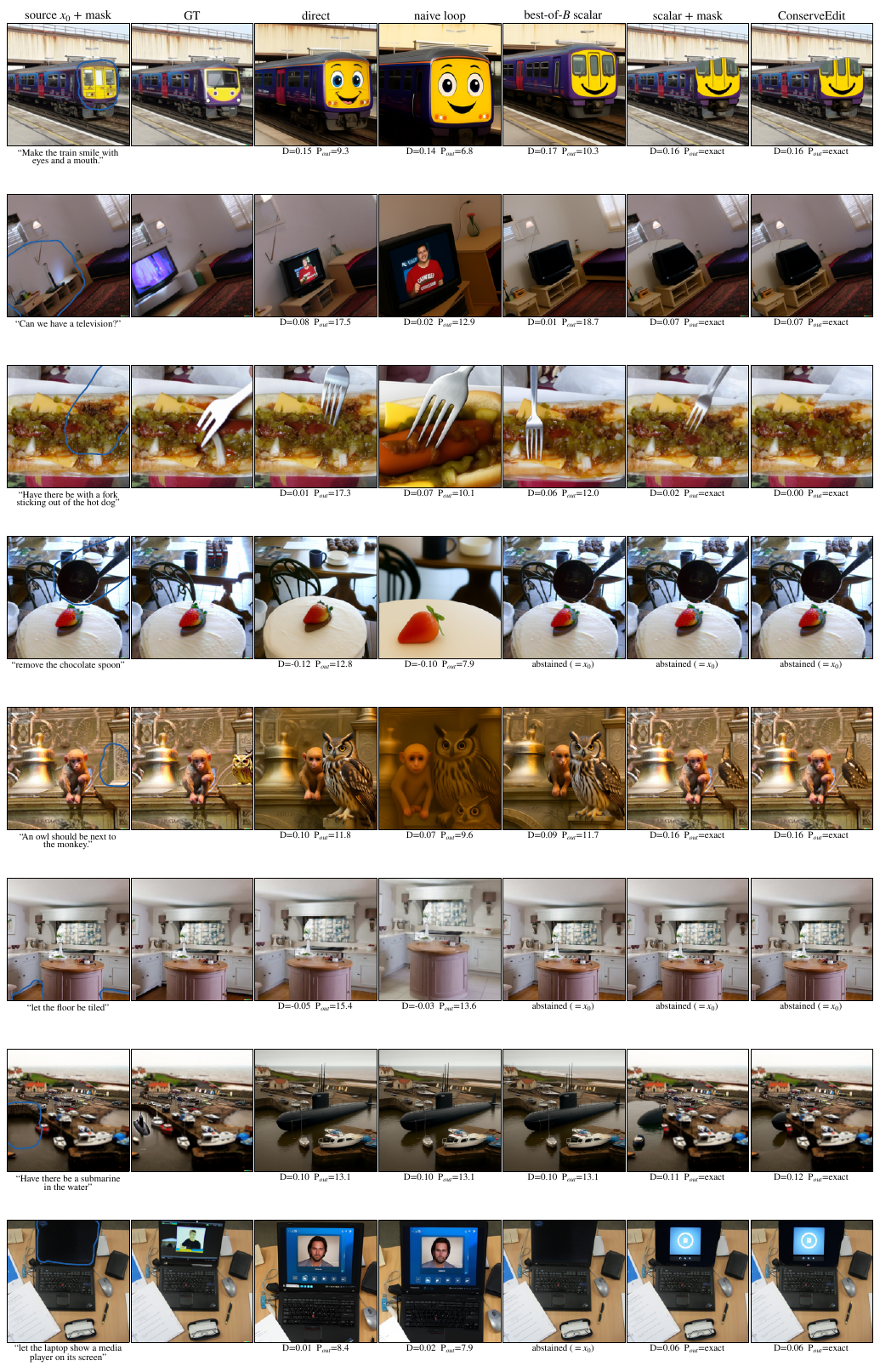}
\caption{\textbf{Additional real Qwen-Image-Edit results, part 2 of 3} (turns 9--16 of the fixed sha1-ordered list; outputs not curated). Captions give the evaluator CLIP-D (D) and PSNR outside the dilated mask (P$_{out}$; ``exact'' means bit-identical to $x_0$).}
\label{fig:s_grid2}
\end{figure*}

\begin{figure*}[p]
\centering
\includegraphics[width=\textwidth,height=0.9\textheight,keepaspectratio]{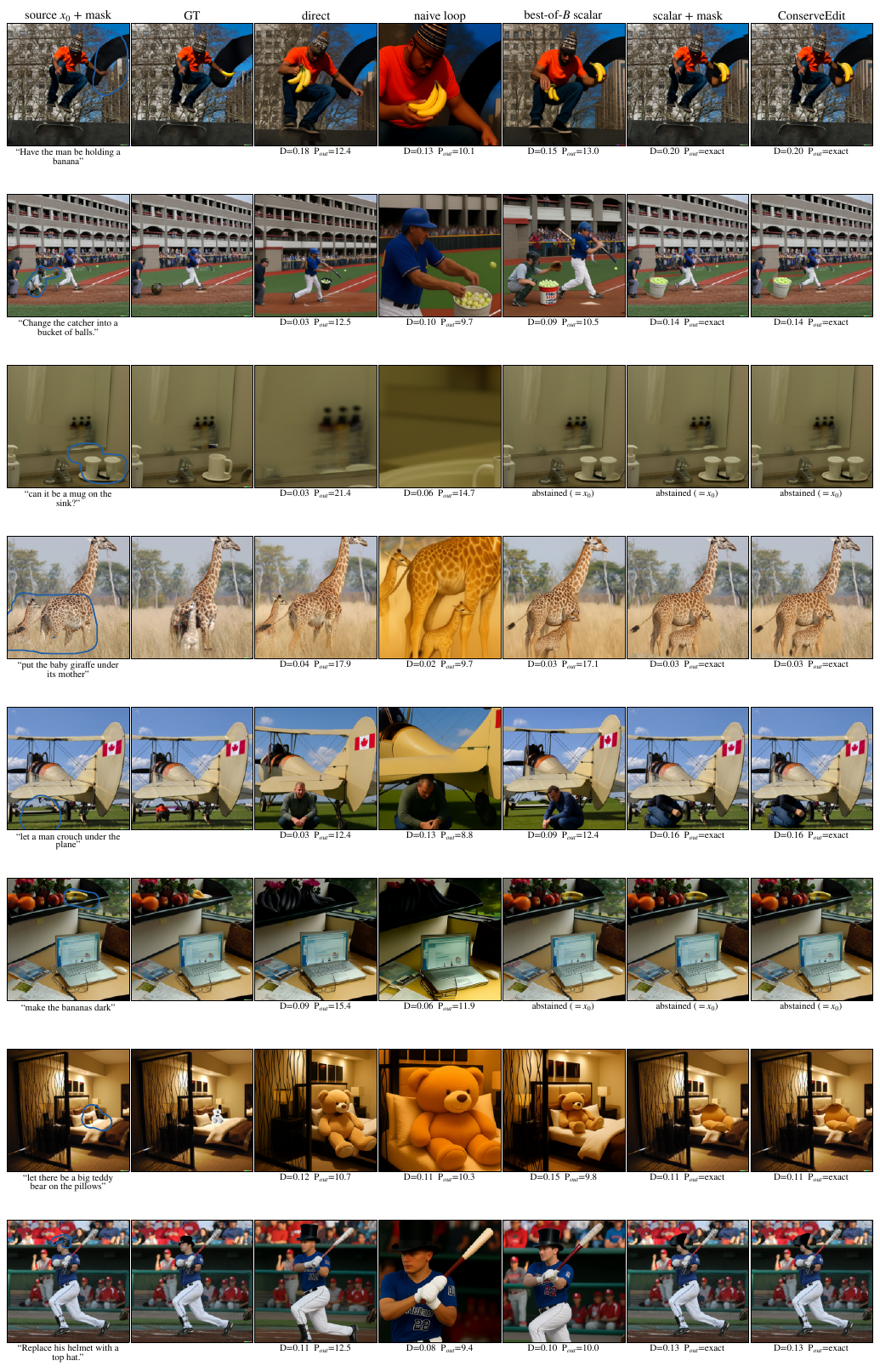}
\caption{\textbf{Additional real Qwen-Image-Edit results, part 3 of 3} (turns 17--24 of the fixed sha1-ordered list; outputs not curated). Captions give the evaluator CLIP-D (D) and PSNR outside the dilated mask (P$_{out}$; ``exact'' means bit-identical to $x_0$).}
\label{fig:s_grid3}
\end{figure*}

\paragraph{More qualitative results.}
\Cref{fig:s_grid1,fig:s_grid2,fig:s_grid3} show 24 further turns. Rule: the 24 turns with the smallest $\mathrm{sha1}(\textit{id})$, excluding the turns of \cref{fig:overview,fig:realqual}; \cref{fig:teaser} shows two of them. The rule does not depend on any output. \method abstains on 5 of these 24 turns. Typical patterns:
\begin{itemize}
\item the naive loop zooms in and re-renders scenes with every iteration;
\item best-of-$B$ scalar selection still shifts the whole image;
\item masked commitment preserves the scene but can leave pasted-in objects that are visually inconsistent with their surroundings.
\end{itemize}

\clearpage

%=====================================================================
\section{SimEdit Details and Full Tables}
\label{sup:sim}

\paragraph{Scene generator.} Each $96{\times}96$ scene (\texttt{conserveedit/sim.py}) contains:
\begin{itemize}
\item a background: a random two-color linear gradient plus a low-frequency sinusoidal texture of amplitude 0.04;
\item the target: a rectangle or ellipse 16--29\,px in size, in a random non-red color;
\item a cast shadow: the target shifted by 2--4\,px down and to the right, darkened by 20\%;
\item a face-like disk of radius 7--10\,px with a diagonal texture, placed 1--2\,px from the target in half of the scenes;
\item a high-frequency striped ``text'' block of $8{\times}(16\text{--}25)$\,px.
\end{itemize}
The controller sees the target only through a noisy grounding: a shift of up to $\pm4$\,px followed by dilation or erosion by $-3$ to $+4$\,px. Over the 200 evaluation scenes this gives a mean IoU of 0.61 and a minimum of 0.30.

\paragraph{Editor model.} \Cref{tab:s_editor} lists the drift processes of the stochastic editor. All of them are applied independently on every call, and all are released as parameters of \texttt{SimEditorConfig}.

\begin{table}[t]
\centering
\caption{\textbf{SimEdit editor model} (default values; \cref{fig:lam}b sweeps $\sigma_{\max}$).}
\label{tab:s_editor}
\scriptsize
\setlength{\tabcolsep}{3pt}
\resizebox{\columnwidth}{!}{%
\begin{tabular}{@{}lll@{}}
\toprule
Process & Effect & Parameter \\
\midrule
Edit & blend target toward red with strength $s\sim\mathcal U[0.35,0.95]$ & fail w.p.\ 0.15 \\
Spill & edit region = target dilated by $k\sim\mathcal U\{0..3\}$\,px & $k_{\max}=3$ \\
Shadow & darken true shadow by $0.1\,s$ & -- \\
Re-synthesis noise & i.i.d.\ Gaussian, $\sigma\sim\mathcal U[0,\sigma_{\max}]$ & $\sigma_{\max}=0.03$ \\
Color cast & global offset $\mathcal N(0,0.03^2)$ per channel & $p=0.30$ \\
Geometric drift & whole-frame shift by $\pm1$--$2$\,px & $p=0.20$ \\
Identity drift & face color offset $\mathcal N(0,0.06^2)$ & $p=0.15$ \\
\bottomrule
\end{tabular}}
\end{table}

\paragraph{Evaluation.} Ground-truth masks are hidden from every controller. The background is the image outside a 3\,px dilation of the true target and shadow, minus the two entities. Constrained success requires $f\ge0.7$, face and text MAE $\le0.01$, and background MAE $\le0.01$.

\paragraph{Full results and tests.} \Cref{tab:s_simfull} gives the full results of all eight controllers (mean $\pm$ SD over scenes, plus background PSNR). \Cref{tab:s_simtests} gives the paired sign-flip permutation tests against the full method ($10^4$ permutations, scene-level means over three seeds).

\begin{table*}[t]
\centering
\caption{\textbf{SimEdit, full results} (200 scenes $\times$ 3 seeds; mean $\pm$ SD over scenes).}
\label{tab:s_simfull}
\small
\begin{tabular}{@{}lcccccc@{}}
\toprule
Controller & CSR (\%) & Target $f$ & BG PSNR (dB) & Ent.\ drift ($10^{-3}$) & DriftAUC ($10^{-3}$) & Calls \\
\midrule
Naive loop & 5.8$\pm$12.7 & 0.873 & 29.5 & 165.83$\pm$111.97 & 187.00 & 2.2 \\
Best-of-$B$ scalar & 14.5$\pm$19.4 & 0.858 & 34.9 & 28.54$\pm$23.91 & 45.21 & 5.2 \\
Scalar\,+\,mask & 55.0$\pm$38.2 & 0.752 & 56.9 & 2.11$\pm$5.98 & 2.15 & 6.5 \\
Gate, current ref. & 50.7$\pm$31.7 & 0.591 & 48.1 & 4.26$\pm$1.65 & 6.09 & 11.9 \\
Gate, no mask & 46.0$\pm$31.1 & 0.552 & 48.5 & 3.99$\pm$1.51 & 5.86 & 12.2 \\
\method w/ current ref. & 61.3$\pm$38.6 & 0.753 & 57.1 & 0.55$\pm$1.33 & 0.61 & 6.8 \\
\method w/o repair & 60.2$\pm$38.9 & 0.742 & 57.2 & 0.65$\pm$1.54 & 0.67 & 6.9 \\
\method (full) & 61.8$\pm$38.3 & 0.752 & 57.1 & 0.51$\pm$1.22 & 0.58 & 6.8 \\
\bottomrule
\end{tabular}

\end{table*}

\begin{table}[t]
\centering
\caption{\textbf{SimEdit, paired tests against \method (full).} $\Delta$CSR = full minus row.}
\label{tab:s_simtests}
\scriptsize
\begin{tabular}{@{}lccc@{}}
\toprule
Controller vs.\ \method & $\Delta$CSR (pts) & $p$ & $d_z$ \\
\midrule
Naive loop & +56.0 & $<10^{-3}$ & 1.39 \\
Best-of-$B$ scalar & +47.3 & $<10^{-3}$ & 1.15 \\
Scalar\,+\,mask & +6.8 & $<10^{-3}$ & 0.33 \\
Gate, current ref. & +11.2 & 0.002 & 0.23 \\
Gate, no mask & +15.8 & $<10^{-3}$ & 0.33 \\
\method w/ current ref. & +0.5 & 0.249 & 0.12 \\
\method w/o repair & +1.7 & 0.033 & 0.16 \\
\bottomrule
\end{tabular}

\end{table}

\paragraph{Success by grounding quality.} \Cref{tab:s_simiou} stratifies constrained success by the IoU of the noisy grounding. Every masked controller fails on all 12 scenes with IoU${<}0.45$: when the support covers too little of the object, no feasible commit can satisfy $\tau$. Unmasked gating succeeds on a third of them, at the cost of exposure to leakage. Above IoU 0.6, \method reaches 78--79\% success. This quantifies the grounding bottleneck discussed in \cref{sec:limits}.

\begin{table*}[t]
\centering
\caption{\textbf{SimEdit constrained success (\%) by grounding IoU} (number of scenes per bin in the second row).}
\label{tab:s_simiou}
\small
\begin{tabular}{@{}lccccc@{}}
\toprule
Controller & IoU$\in[0.00,0.45)$ & IoU$\in[0.45,0.60)$ & IoU$\in[0.60,0.75)$ & IoU$\in[0.75,1.00)$ & all \\
 & $n{=}12$ & $n{=}78$ & $n{=}91$ & $n{=}19$ & $n{=}200$ \\
\midrule
Naive loop & 2.8 & 6.4 & 5.9 & 5.3 & 5.8 \\
Best-of-$B$ scalar & 5.6 & 16.2 & 14.3 & 14.0 & 14.5 \\
Scalar\,+\,mask & 0.0 & 41.9 & 70.0 & 71.9 & 55.0 \\
Gate, current ref. & 33.3 & 49.6 & 55.3 & 43.9 & 50.7 \\
Gate, no mask & 33.3 & 45.7 & 49.1 & 40.4 & 46.0 \\
\method w/ current ref. & 0.0 & 47.9 & 77.3 & 78.9 & 61.3 \\
\method w/o repair & 0.0 & 45.3 & 76.9 & 78.9 & 60.2 \\
\method (full) & 0.0 & 48.3 & 78.0 & 78.9 & 61.8 \\
\bottomrule
\end{tabular}

\end{table*}

\paragraph{Repair and reference image.} Disabling repair costs 1.7 points of CSR ($p{=}0.03$). Without a mask, a current-image reference accepts more edits than the source reference (50.7\% vs.\ 46.0\% CSR) but lets drift grow with the number of calls (\cref{fig:traj}a), as \cref{prop:goal} predicts. With a mask, the two references differ little (61.3\% vs.\ 61.8\%, $p{=}0.25$; drift difference $p{=}0.015$), because the mask already confines drift to the overlap between support and entities.

\paragraph{Matched editor-call budgets.} At 4, 8, 16, and 32 calls, \method reaches 31.0, 53.0, 55.7, and 55.7\% CSR, compared with 30.3, 49.7, 50.0, and 50.0\% for Scalar\,+\,mask and about 20\% for unmasked best-of-$B$ (\cref{fig:traj}b). Extra samples do not substitute for the acceptance logic, and returns saturate at about 16 calls.

\begin{figure*}[t]
\centering
\includegraphics[width=\textwidth]{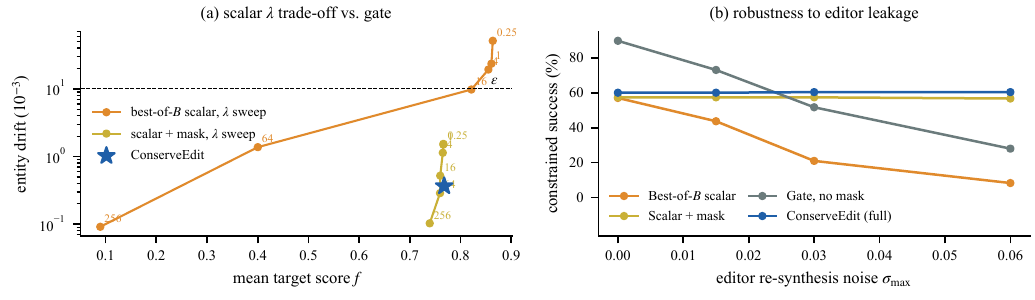}
\caption{\textbf{(a)} Sweeping the scalar weight $\lambda$ (labels). Without a mask, no $\lambda$ reaches low drift without collapsing the target score (\cref{prop:scalar}). With a mask, a large $\lambda$ approaches \method (star) but must be tuned. \textbf{(b)} Constrained success as the editor's re-synthesis noise grows (100 scenes $\times$ 3 seeds). Unmasked gating is best for a leak-free editor but collapses under leakage; masked controllers stay flat.}
\label{fig:lam}
\end{figure*}

\paragraph{Scalar weight.} Without a mask, the $\lambda$ sweep in \cref{fig:lam}a traces the trade-off of \cref{prop:scalar}: $\lambda{=}16$ keeps $f{=}0.82$ but sits at the drift tolerance (44.3\% CSR), whereas $\lambda{=}64$ lowers drift but collapses the target ($f{=}0.40$). With a mask, the scalar loop approaches \method at large $\lambda$ (63.3\% at $\lambda{=}64$ vs.\ 64.0\% on the same scenes); \method reaches this level without a weight to tune and logs every decision.

\paragraph{Editor leakage and grounding.} With a leak-free editor, unmasked gating reaches 89.7\% versus 60.0\% for \method (\cref{fig:lam}b), because a mask that misses part of the object caps achievable progress. Under realistic leakage the order reverses (28.0\% vs.\ 60.3\% at $\sigma_{\max}{=}0.06$). Masked commitment thus trades sensitivity to grounding for robustness to leakage (\cref{tab:s_simiou}).

\paragraph{Sweeps.} \Cref{tab:s_budget,tab:s_lambda,tab:s_stress} give the exact values behind \cref{fig:traj}b and \cref{fig:lam} (100 held-out scenes $\times$ 3 seeds each).

\begin{figure*}[t]
\centering
\includegraphics[width=\textwidth]{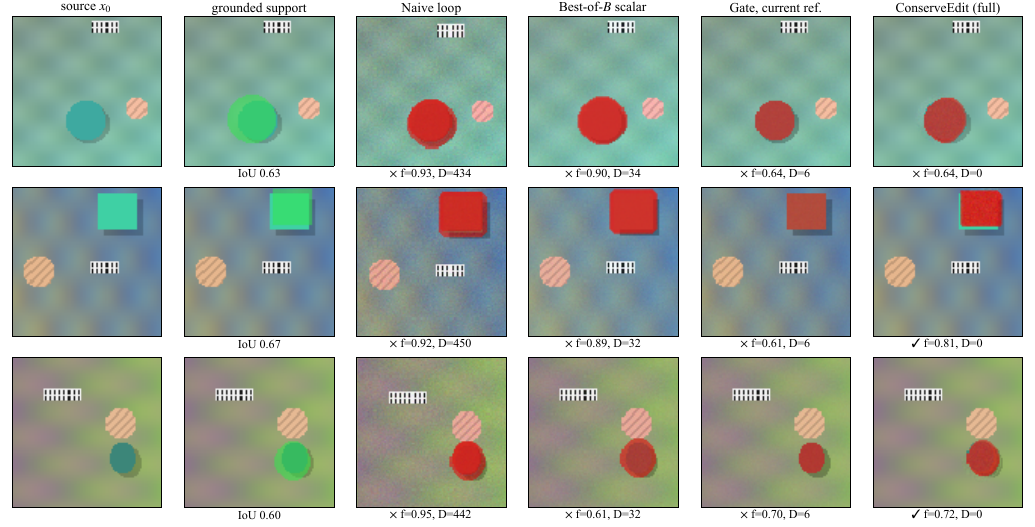}
\caption{\textbf{SimEdit qualitative results} (rule: the first three scenes of a held-out seed range with grounding IoU in $(0.45,0.8)$ that contain a face). The second column shows the grounded support; magenta marks overlap with the face. Captions give CSR (\ding{51}/\ding{55}), target score $f$, and entity drift $D$ in $10^{-3}$.}
\label{fig:s_simqual}
\end{figure*}

\paragraph{Qualitative examples.} \Cref{fig:s_simqual} shows three held-out scenes in which the grounded support overlaps the face. The naive loop meets the target in all three but with entity drift above $0.4$, more than $40\epsilon$; best-of-$B$ scalar selection still exceeds $\epsilon$ about threefold. \method keeps entity drift below $0.5\times10^{-3}$ (shown as $D{=}0$) and succeeds in two of the three scenes; in the first it stops below $\tau$ ($f{=}0.64$).

\begin{table}[t]
\centering
\caption{\textbf{Budget sweep:} constrained success (\%) vs.\ editor-call budget ($B{=}4$, $T{=}1,2,4,8$).}
\label{tab:s_budget}
\scriptsize
\begin{tabular}{@{}lcccc@{}}
\toprule
Editor-call budget & 4 & 8 & 16 & 32 \\
\midrule
Naive loop & 7.3 & 7.3 & 7.3 & 7.3 \\
Best-of-$B$ scalar & 18.7 & 19.3 & 20.0 & 20.0 \\
Scalar\,+\,mask & 30.3 & 49.7 & 50.0 & 50.0 \\
\method (full) & 31.0 & 53.0 & 55.7 & 55.7 \\
\bottomrule
\end{tabular}

\end{table}

\begin{table}[t]
\centering
\caption{\textbf{Scalar-weight sweep.} CSR (\%), mean target score $f$, and entity drift of both scalar loops; \method uses no weight.}
\label{tab:s_lambda}
\scriptsize
\setlength{\tabcolsep}{2.6pt}
\begin{tabular}{@{}lcccccc@{}}
\toprule
& \multicolumn{3}{c}{Best-of-$B$ scalar} & \multicolumn{3}{c}{Scalar\,+\,mask} \\
\cmidrule(lr){2-4}\cmidrule(l){5-7} $\lambda$ & CSR & $f$ & Ent.\ ($10^{-3}$) & CSR & $f$ & Ent.\ ($10^{-3}$) \\
\midrule
0.25 & 20.0 & 0.86 & 51.33 & 61.3 & 0.77 & 1.54 \\
1 & 22.7 & 0.86 & 23.54 & 61.3 & 0.77 & 1.50 \\
4 & 27.3 & 0.85 & 19.30 & 62.0 & 0.76 & 1.14 \\
16 & 44.3 & 0.82 & 9.78 & 62.7 & 0.76 & 0.52 \\
64 & 32.0 & 0.40 & 1.38 & 63.3 & 0.76 & 0.29 \\
256 & 6.7 & 0.09 & 0.09 & 62.0 & 0.74 & 0.10 \\
\midrule
\method & \multicolumn{3}{c}{--} & 64.0 & 0.77 & 0.37 \\
\bottomrule
\end{tabular}

\end{table}

\begin{table}[t]
\centering
\caption{\textbf{Editor-leakage sweep:} constrained success (\%) vs.\ maximum re-synthesis noise $\sigma_{\max}$.}
\label{tab:s_stress}
\scriptsize
\begin{tabular}{@{}lcccc@{}}
\toprule
$\sigma_{\max}$ & 0 & 0.015 & 0.03 & 0.06 \\
\midrule
Best-of-$B$ scalar & 57.0 & 43.7 & 21.0 & 8.3 \\
Scalar\,+\,mask & 57.3 & 57.3 & 57.3 & 56.7 \\
Gate, no mask & 89.7 & 73.0 & 51.7 & 28.0 \\
\method (full) & 60.0 & 60.0 & 60.3 & 60.3 \\
\bottomrule
\end{tabular}

\end{table}

%=====================================================================
\section{Complete Proofs and Additional Guarantees}
\label{sup:theory}

\paragraph{Setting.} Images are $x\in[0,1]^{H\times W\times 3}$; supports are subsets of the pixel grid $\Omega$. At step $t$ the controller selects a clause $j_t$ with current support $A_{j_t}^{(t)}$ and a commit mask $W_t:\Omega\to[0,1]$ with $\mathrm{supp}(W_t)\subseteq A_{j_t}^{(t)}$; the implementation feathers the binary support and multiplies by it. The cumulative support is $U_t=\bigcup_{\ell<t,\ \ell\text{ committed}} A_{j_\ell}^{(\ell)}$. The executor $E_\phi$ is arbitrary, possibly adversarial; no property of it is assumed.

\paragraph{Proofs of the main-paper propositions.}

\begin{proof}[Proof of \cref{prop:inv}]
\emph{Invariance.} By induction on $t$. For $t{=}0$, $x_0{=}x_0$ and $U_0{=}\emptyset$. Assume $x_t(p){=}x_0(p)$ for all $p\notin U_t$. If step $t$ rolls back, $x_{t+1}{=}x_t$ and $U_{t+1}{=}U_t$. If it commits, then for $p\notin A_{j_t}^{(t)}$ we have $W_t(p){=}0$, so $x_{t+1}(p)=x_t(p)$ by \cref{eq:compose}. Hence every $p\notin U_{t+1}=U_t\cup A_{j_t}^{(t)}$ satisfies $x_{t+1}(p)=x_t(p)=x_0(p)$.
\emph{Drift bound.} Every committed state passed \cref{eq:c1} with descriptors computed from $x_0$ once, before the loop; every other state equals the last committed state or $x_0$, and $D_k(x_0,x_0)=0$. Hence $D_k(x_0,x_t)\le\epsilon_k$ for all $t$, which bounds both the maximum and the average over the trajectory. The bound does not involve $T$ or $B$.
\end{proof}

\begin{proof}[Proof of \cref{prop:scalar}]
Fix $x_t$ with $D(x_0,x_t)=0$ and $f(x_t)=f_t<1-\delta$. Consider two candidates: $a$ with $(\Delta f,D)=(\delta,\epsilon)$, which is feasible, and $b_\eta$ with $(\Delta f,D)=(1-f_t,\epsilon+\eta)$, which is infeasible for every $\eta>0$. The scalar rule accepts a candidate iff $\Delta f-\lambda D>0$.
If $\lambda<(1-f_t)/\epsilon$, choose $\eta<(1-f_t)/\lambda-\epsilon$; then $b_\eta$ is accepted.
If $\lambda\ge\delta/\epsilon$, then $\delta-\lambda\epsilon\le0$ and $a$ is rejected.
Since $\delta/\epsilon<(1-f_t)/\epsilon$, every $\lambda>0$ satisfies at least one of the two conditions.
\end{proof}

\paragraph{Additional guarantees.}

\begin{proposition}[Goalpost drift]\label{prop:goal}
Let $d$ be any metric on entity content (\eg, masked mean absolute error). If the audit instead compares a candidate with the \emph{current} image, $d(x_t,x_{t+1})\le\epsilon$, then $d(x_0,x_T)\le T\epsilon$, and this bound is tight: there exist trajectories that pass every per-step audit with $d(x_0,x_T)=T\epsilon$ (up to the diameter of the space).
\end{proposition}
\begin{proof}
The upper bound follows from the triangle inequality, $d(x_0,x_T)\le\sum_{t<T}d(x_t,x_{t+1})\le T\epsilon$. Tightness: let $d$ be the mean absolute error over an entity of $n$ pixels, and let every candidate add $\epsilon$ to every entity pixel, staying inside $[0,1]$ as long as $T\epsilon\le 1-\max x_0$. Each step has $d(x_t,x_{t+1})=\epsilon$ and passes the current-reference audit, while $d(x_0,x_T)=T\epsilon$, because all increments share one sign.
\end{proof}

\begin{proposition}[Termination]\label{prop:term}
For a single clause with $\delta>0$, at most $\lceil(\tau-f(x_0))/\delta\rceil$ commits occur before the clause is satisfied; the loop terminates after at most $T$ iterations and $TB$ editor calls.
\end{proposition}
\begin{proof}
For a single clause, every commit raises $f$ by at least $\delta$ (\cref{eq:c2}), so after $c$ commits $f\ge f(x_0)+c\delta$; the clause is met once $c\ge\lceil(\tau-f(x_0))/\delta\rceil$. Independently, the loop executes at most $T$ iterations with at most $B$ executor calls each.
\end{proof}

\begin{proposition}[Repairs never authorize protected content]\label{prop:s_repair}
If $A_j^{(0)}\cap P_k=\emptyset$ for all protected entities $P_k$, this remains true after any sequence of repairs. Without that assumption (the non-strict contracts used with noisy grounding), shrink repairs only remove protected pixels from the support, and halo expansion never adds any.
\end{proposition}
\begin{proof}
Shrink sets $A\leftarrow A\setminus P_k$. Halo expansion sets $A\leftarrow\mathrm{dilate}(A,1)\setminus\bigcup_k P_k$. Neither operation adds a protected pixel to $A$, and seed resampling leaves $A$ unchanged.
\end{proof}

\begin{proposition}[Abstention is exact]\label{prop:s_abstain}
If no transaction is committed, the output equals $x_0$ bit for bit, and the decision log records the violated clauses of every rejected candidate.
\end{proposition}
\begin{proof}
The state changes only on commit (\cref{alg:loop}), and every evaluated candidate is appended to the decision log with its violations before the acceptance decision.
\end{proof}

\paragraph{What the guarantees do not cover.} All statements refer to the \emph{represented} contract.
(a)~If grounding misses part of the target, \cref{prop:inv} still holds, but the edit may be incomplete (\cref{tab:s_simiou}).
(b)~If the descriptors are blind to a change (\eg, an identity shift that preserves pixel statistics), \cref{prop:inv} certifies an incorrect state.
(c)~Nothing bounds the quality of the content \emph{inside} $U_T$; seams and implausible composites (\cref{fig:s_seams}) are outside the guarantees.
(d)~A false rejection by the progress predicate (\cref{fig:s_abstain}) costs utility, not safety (\cref{prop:s_abstain}).

%=====================================================================
\section{Reproducibility, Assets, and Licenses}
\label{sup:repro}

\paragraph{Released artifacts.} The anonymized code release contains:
\begin{itemize}\raggedright
\item the controller (\texttt{conserveedit/}: contract, audits, transactional loop, repairs, decision log);
\item the SimEdit environment and driver (\texttt{experiments/run\_sim.py});
\item the real-editor pipeline (\texttt{run\_qwen.py}, \texttt{eval\_real.py}, \texttt{analyze\_real.py}), the Direct\,+\,mask baseline (\texttt{direct\_mask.py}), automatic grounding (\texttt{auto\_ground.py}), and the inpainting baseline (\texttt{fill\_baseline.py}), the face/text study (\texttt{entities.py}, \texttt{entity\_prep.py}, \texttt{entity\_eval.py}), the predicate study (\texttt{predicates.py}, \texttt{calib\_pred.py}, \texttt{pred\_eval.py}), the resolution study (\texttt{res\_eval.py}), the agentic-editor comparison (\texttt{agents\_run.py}, \texttt{er\_run.py}, \texttt{reasoning\_infer.py}, \texttt{er\_compat/}, \texttt{agents\_eval.py}), and the paired tests with Holm correction (\texttt{stats.py}, \texttt{analyze\_v2.py});
\item the figure and supplementary scripts (\texttt{fig\_real.py}, \texttt{fig1\_assets.py}, \texttt{make\_teaser.py}, \texttt{make\_teaser\_entities.py}, \texttt{gallery\_extract.py}, \texttt{suppl\_real.py}, \texttt{suppl\_sim.py});
\item unit tests;
\item for all 1{,}053 real turns, every controller's output image and the full per-turn decision log (accepted and rejected transactions, violations, repairs, editor calls);
\item per-sample metrics and the exact environment: Python 3.11, torch 2.11.0+cu128, diffusers 0.42.1, transformers 5.22.1, with a lock file of all 76 packages and a one-step setup script for a fresh GPU server; the agentic baselines add peft 0.21.3 and pandas 3.1.1, and RePlan and EditRefiner load diffusers 0.37.1 from a separate path.
\end{itemize}

\paragraph{Reproducing the numbers.}
\begin{itemize}\raggedright
\item \cref{fig:gate,fig:traj,fig:lam}, plus \cref{tab:s_simfull,tab:s_simtests,tab:s_simiou,tab:s_budget,tab:s_lambda,tab:s_stress}: \texttt{python experiments/run\_sim.py} (deterministic; CPU, 170\,s) followed by \texttt{suppl\_sim.py}.
\item \cref{tab:real} and \cref{fig:realqual}: \texttt{make\_full\_list.py}, then \texttt{run\_qwen.py -{}-ids\_file lists/test\_full.txt} with all six controllers and Direct\,+\,mask (16 shards on 8 GPUs; every turn is written to disk, so runs resume), then \texttt{eval\_real.py}, \texttt{analyze\_real.py}, \texttt{analyze\_v2.py exp1}, \texttt{make\_table\_real.py}, and \texttt{fig\_real.py}. \texttt{pipeline\_exp3.sh} chains these steps.
\item the FLUX.1 Fill rows of \cref{tab:real,tab:auto}: \texttt{fill\_baseline.py} (human and automatic masks), \texttt{eval\_real.py -{}-methods fill,fill\_mask}, \texttt{analyze\_v2.py fill}.
\item \cref{tab:entities}: \texttt{entity\_prep.py select} (test sources) and \texttt{calibrate} (dev sources), \texttt{run\_qwen.py -{}-entities\_dir \ldots\ -{}-entity\_tol \ldots} with human and automatic masks, \texttt{eval\_real.py}, \texttt{entity\_eval.py}.
\item \cref{tab:s_pred}: \texttt{calib\_pred.py} (dev), \texttt{run\_qwen.py -{}-pred\_calib \ldots\ [-{}-pred\_noseam]}, \texttt{eval\_real.py}, \texttt{pred\_eval.py}, \texttt{make\_table\_pred.py}.
\item \cref{tab:s_agents} (\cref{sup:agents}): \texttt{agents\_run.py -{}-method editthinker} and \texttt{-{}-method replan}, \texttt{er\_run.py} (EditRefiner, run with \texttt{er\_compat/} on the Python path), each with \texttt{-{}-ids\_file lists/subset160.txt}; then \texttt{eval\_real.py -{}-methods $m$,$m$\_mask}, \texttt{agents\_eval.py}, and \texttt{make\_table\_real.py} with the generated \texttt{table\_agents\_rows.tex}.
\item \cref{tab:s_res}: \texttt{run\_qwen.py -{}-res 1024 -{}-halo 16 -{}-ids\_file lists/subset160.txt}, then \texttt{res\_eval.py view}, \texttt{eval\_real.py}, and \texttt{res\_eval.py table}.
\item \cref{tab:auto}: \texttt{auto\_ground.py} on dev, \texttt{-{}-calibrate}, then on test; \texttt{run\_qwen.py -{}-mask\_source auto}; \texttt{eval\_real.py}; \texttt{analyze\_v2.py exp2}; \texttt{make\_table\_auto.py}.
\item \cref{fig:overview}: \texttt{fig1\_assets.py}, which re-runs one turn with the logged seeds.
\item \cref{fig:teaser}: \texttt{make\_teaser.py}, which takes the two turns of \cref{fig:s_grid1,fig:s_grid2,fig:s_grid3} named in it and recovers their images from the gallery renders (\texttt{gallery\_extract.py}); the right panel: \texttt{make\_teaser\_entities.py}, which reads \cref{tab:entities}.
\item \cref{sup:real}: \texttt{suppl\_real.py}; every selection rule is a constant in that file.
\end{itemize}
All 1{,}053 MagicBrush test turns are used. Runs process them in a fixed random permutation (seed 0) whose first 160 turns form the subset used for the $1024^2$ and agentic-editor comparisons, so every completed prefix is itself a random subset.

\paragraph{Assets and licenses.}
\begin{itemize}\raggedright
\item MagicBrush~\cite{zhang2023magicbrush}: CC-BY-4.0; test split, and dev split for calibrating $\tau_{env}$.
\item Qwen-Image-Edit~\cite{wu2025qwenimage}: Apache-2.0.
\item DINOv2~\cite{oquab2024dinov2}: Apache-2.0.
\item CLIP ViT-B/32 and ViT-L/14~\cite{radford2021clip}: used under the terms of OpenAI's model cards.
\item Grounding DINO-base~\cite{liu2024groundingdino} (\texttt{IDEA-Research/grounding-dino-base}): Apache-2.0.
\item SAM ViT-H~\cite{kirillov2023sam} (\texttt{facebook/sam-vit-huge}): Apache-2.0.
\item insightface~\cite{deng2019arcface} \texttt{buffalo\_l} and \texttt{antelopev2}: code MIT; pretrained models for non-commercial research only.
\item SigLIP so400m~\cite{zhai2023siglip} (\texttt{google/siglip-so400m-patch14-384}): Apache-2.0.
\item EasyOCR: Apache-2.0. TrOCR~\cite{li2023trocr} (\texttt{microsoft/trocr-base-printed}): released with the MIT-licensed \texttt{microsoft/unilm} code; the model card states no license.
\item FLUX.1 Fill [dev]~\cite{bfl2024fill} (\texttt{black-forest-labs/FLUX.1-Fill-dev}): FLUX.1 [dev] Non-Commercial License, used for non-commercial research.
\item spaCy 3.8 and \texttt{en\_core\_web\_sm} 3.8.0 (phrase extraction): MIT.
\item EditThinker~\cite{li2025editthinker}: neither the repository nor the model card states a license; used only for this non-commercial research comparison.
\item RePlan~\cite{qu2026replan}: code and weights Apache-2.0.
\item EditRefiner~\cite{xu2026editrefiner}: code MIT; the model card of its adapters states no license.
\item Qwen3-VL-8B-Instruct~\cite{bai2025qwen3vl} and Qwen-Image-Edit-2511 (used inside the agentic baselines): Apache-2.0.
\end{itemize}
No human subjects were involved.

\end{document}